%% file: iclr2025_conference.tex
\documentclass{article} 
\usepackage{iclr2025_conference,times}

\input{math_commands.tex}

\usepackage{siunitx}

\usepackage{hyperref}
\usepackage{url}

\usepackage[utf8]{inputenc} 
\usepackage[T1]{fontenc}    
\usepackage{amsfonts}       
\usepackage{nicefrac}       
\usepackage{microtype}      
\usepackage{xspace}
\usepackage{dsfont}
\usepackage{times}
\usepackage{latexsym}
\usepackage{balance}
\usepackage{bbm}
\usepackage{paralist}
\usepackage{makecell}
\usepackage{wrapfig}
\usepackage{amsmath}
\usepackage{amssymb}
\usepackage{mathtools}
\usepackage{amsthm}
\usepackage{multicol}
\usepackage{subcaption}
\newtheorem{lemma}{Lemma}[section]
\newtheorem{theorem}{Theorem}[section]
\usepackage{hyperref}
\usepackage{xcolor}
\usepackage{booktabs, multirow}
\usepackage{tabularx,ragged2e} 
\newcolumntype{L}[1]{>{\RaggedRight\hsize=#1\hsize}X}
\newcolumntype{C}[1]{>{\Centering\hsize=#1\hsize}X} 
\usepackage{multibib}
\usepackage{color,soul}
\usepackage{colortbl}
\usepackage{todonotes}
\usepackage{tcolorbox}
\usepackage{graphicx}
\usepackage{stackengine}
\usepackage{enumitem}
\usepackage{listings}
\usepackage{floatflt}

\title{Triple Preference Optimization: Achieving Better Alignment using a Single Step Optimization}

\author{Amir Saeidi$^{1}$, Shivanshu Verma$^{1}$, Aswin RRV$^1$, Kashif Rasul$^2$, Chitta Baral$^1$ \\
$^1$Arizona State University, $^2$Morgan Stanley \\
\texttt{\{ssaeidi1, chitta\}@asu.edu} \\
}

\iclrfinalcopy 

\begin{document}

\maketitle

\input{text/0_abstract}
\input{text/1_introduction}

\input{text/3_method}

\input{text/4_experimental_setup}

\input{text/5_experiment_results}

\input{text/2_related_work}

\input{text/6_conclusion}

\subsubsection*{Acknowledgments}
We thank the Research Computing (RC) at Arizona State University (ASU) and \href{https://www.cr8dl.ai/}{cr8dl.ai} for their generous support in providing computing resources. The views and opinions of the authors expressed herein do not necessarily state or reflect those of the funding agencies and employers. We acknowledge support by a 2023 Spring Amazon Research Award (ARA).
\bibliography{anthology}
\bibliographystyle{iclr2025_conference}
\input{text/7_appendix}

\end{document}

%% file: math_commands.tex
\usepackage{amsmath,amsfonts,bm}

\def\eqref#1{equation~\ref{#1}}

\def\1{\bm{1}}

\DeclareMathAlphabet{\mathsfit}{\encodingdefault}{\sfdefault}{m}{sl}
\SetMathAlphabet{\mathsfit}{bold}{\encodingdefault}{\sfdefault}{bx}{n}



%% file: text/0_abstract.tex
\begin{abstract}
Reinforcement Learning with Human Feedback (RLHF) enhances the alignment of Large Language Models (LLMs). However, its limitations have led to the development of Direct Preference Optimization (DPO), an RL-free approach designed to overcome these shortcomings. While studies have shown that DPO improves instruction-following capabilities, it negatively impacts the reasoning ability of LLMs. Additionally, DPO is highly sensitive to judgment noise in preference datasets and the size of the training set. Although several modifications to DPO have been proposed, they still fail to fully resolve these issues.  
To address these limitations, we propose \textbf{Triple Preference Optimization (TPO)}, a new preference learning method designed to enhance both reasoning and instruction-following abilities through one-step optimization. We compare TPO against DPO and its recent variants using state-of-the-art training setups, including both base and instruction-tuned models such as Mistral and Llama~3. Our evaluation covers a comprehensive range of chat-based and reasoning benchmarks.  
The results demonstrate that TPO achieves significant improvements over existing methods without substantially increasing response length across different dataset sizes. Specifically, TPO outperforms DPO and SimPO by up to 7.0\% and 7.3\% points on Arena-Hard, 12.2\% and 13.3\% points on MixEval-Hard, 10.4\% and 10.1\% points on MMLU-Pro, and 19.0\% and 19.2\% points on GSM8K, respectively. Furthermore, TPO achieves these improvements while requiring less data than DPO. \footnote{Code and models can be found at \url{https://github.com/sahsaeedi/TPO/tree/main}}
\end{abstract}

%% file: text/1_introduction.tex
\section{Introduction}
Large language models (LLMs) have recently demonstrated remarkable performance across various tasks \citep{bubeck2023sparks}. While supervised fine-tuning with instruction-based data has significantly enhanced their capabilities, obtaining high-quality human-generated data for new tasks remains a significant challenge \citep{sanh2021multitask, touvron2023llama}. To address this issue, Reinforcement Learning with Human Feedback (RLHF) has been introduced, enabling models to learn from preference data that captures the human perspective \citep{christiano2017deep, stiennon2020learning, ouyang2022training}. To address challenges such as reward hacking in RLHF and its heavy reliance on an explicit reward model, the Direct Preference Optimization (DPO) \citep{rafailov2024direct} method was proposed as an RL-free optimization approach. DPO, unlike RLHF, optimizes a policy model using an implicit reward function where the regular KL divergence acts as the reward.

Recent studies \citep{tunstall2023zephyrDD, xu2024contrastive} indicate that DPO faces challenges related to optimization inefficiency and the need for multi-step optimization. Additionally, new findings suggest that KL divergence may not effectively represent a reward, with the average likelihood of a response given the input offering a potentially better approach for reward modeling \citep{meng2024simpo}. Motivated by these issues, several modified versions of DPO have been proposed \citep{Azar2023AGT, Ethayarajh2024KTOMA, hong2024reference}. As highlighted in \citep{wu2024thinking, meng2024simpo}, preference optimization significantly enhances performance in instruction-following tasks, leading to a notable decline in performance on reasoning benchmarks such as GSM8K. Furthermore, as noted in \citep{saeidi2024insights}, our analysis indicates that current preference optimization methods are highly sensitive to the size of the training dataset.

Motivated by the challenges in current methods, we propose \textbf{Triple Preference Optimization (TPO)}, a novel preference optimization algorithm designed to achieve impressive performance on both instruction-following and reasoning benchmarks in a single-step optimization process. TPO optimizes a pre-trained model by maximizing the likelihood of \( y_{\text{gold}} \) using a behavioral cloning (BC) objective while incorporating a preference optimization term in a reference-free format. This term increases the likelihood of preferred responses(\( y_w \)) and decreases the likelihood of rejected responses (\( y_l \)). Moreover, to enhance control over response length, inspired by the SimPO method, we introduce \textbf{TPO-L}, a length-controlled variant of TPO. In TPO-L, the average length of preferred and rejected responses is used to constrain the policy model, ensuring that it generates longer responses only when it improves quality.

\begin{figure}[t]
    \centering
    \includegraphics[width=\linewidth]{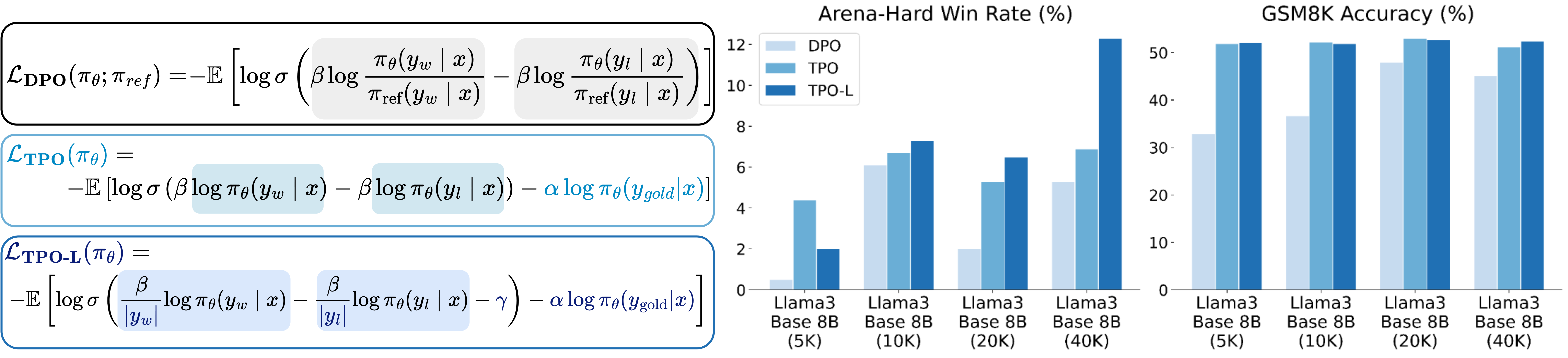}
    \caption{TPO and TPO-L differ by removing the reference model and adding behavioral cloning objective with a regularization term for gold preferences, distinct from preferred and rejected responses. TPO and TPO-L outperform DPO in instruction following and reasoning benchmarks simultaneously.}
    \label{fig:overview_tpo_loss}
    
\end{figure}

We provide a theoretical formulation of TPO by deriving the optimization function of TPO from Maximum Entropy Reinforcement Learning (MERL) by considering a BC objective on gold response as an entropy. We experimentally demonstrate that TPO addresses the optimization conflict challenge present in some recent preference optimization methods. Additionally, we show that using the average likelihood of a sequence of responses provides a more effective representation of the implicit reward than the definitions used in DPO and SimPO. Comprehensive experiments demonstrate that TPO and TPO-L achieve simultaneous improvements in instruction-following and reasoning benchmarks, demonstrating greater robustness across varying training dataset sizes. Furthermore, we observe that TPO exhibits stronger resilience to judgment noise in preference datasets compared to DPO. Notably, even with half the amount of data, TPO and TPO-L consistently outperform DPO. 

In summary, our contributions are as follows:

\begin{itemize}

    \item We propose Triple Preference Optimization (TPO) and TPO-L, novel preference learning methods designed to optimize a policy model in a single step (See Figure \ref{fig:overview_tpo_loss}), achieving significant performance on smaller training datasets compared to existing methods.
    
    \item Extensive experiments demonstrate that TPO and TPO-L significantly outperform existing preference optimization methods on both instruction-following and reasoning benchmarks across various training dataset sizes (See Tables \ref{tab:main_res} and \ref{tab:main_instruct_res}).
    
    \item Theoretically, we show that \( \log \pi_\theta (y|x) \) as an implicit reward in the context of TPO and TPO-L, and comprehensive experiments confirm that it significantly enhances reward modeling compared to DPO and SimPO (See Figures \ref{fig:tpo_reward_modelling} and \ref{fig:reward_modelleing_tpo_dpo}).

    \item We also illustrate that TPO overcomes the conflict optimization challenges in current preference optimization methods (See Figure \ref{fig:conflict_optimization}), exhibits greater robustness to judgment noise in data (See Figure \ref{fig:noise_analysis}), and achieves comparable performance while utilizing less data compared to DPO (See Table \ref{tab:sft-analysis-dpo}).
    
\end{itemize}

%% file: text/3_method.tex
\section{Method}
In this section, we start by discussing the ongoing challenges in current preference optimization methods, which motivate the development of a new method. We then introduce the Triple Preference Optimization (TPO) method, a novel preference learning algorithm, and provide a detailed explanation of the theoretical foundations supporting it.

\subsection{Challenges}
\label{sec:method_challenges}

\begin{figure}
    \centering
    \includegraphics[width=1\linewidth]{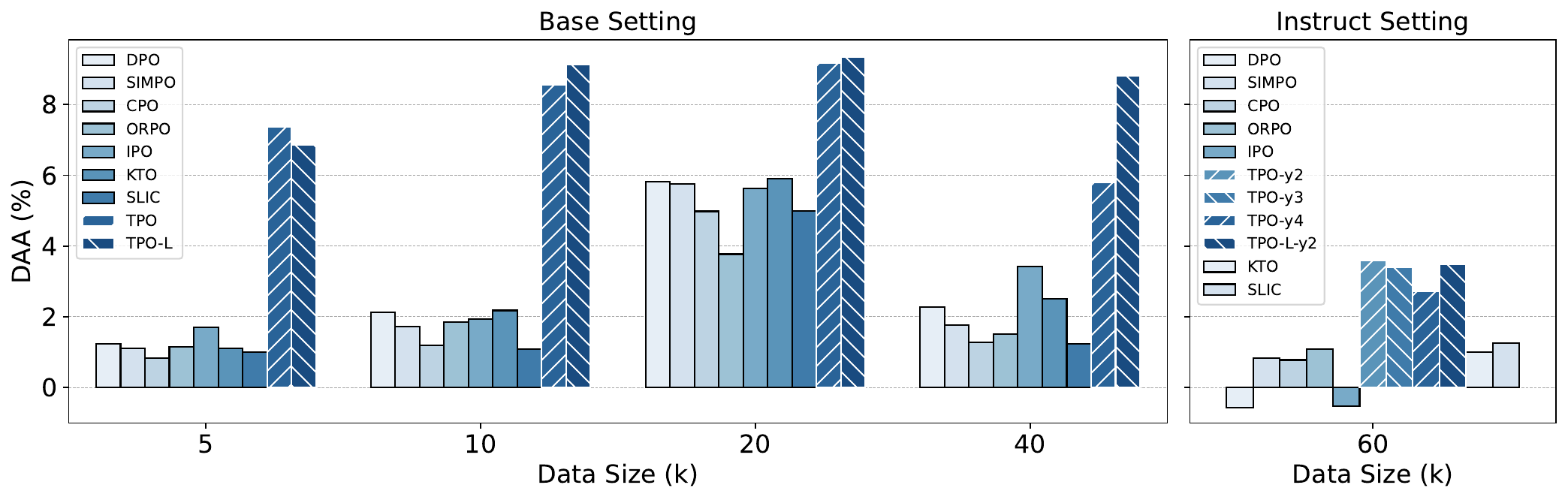}
    \caption{Comparison of improvements achieved during the post-training stage, as measured by the DAA metric, by evaluating the performance of the SFT checkpoint against the preference optimization checkpoint on downstream tasks (More details in Appendix \ref{sec:app_down_stream_tasks}).}
    \label{fig:analysis_sensitivity}
\end{figure}

Current preference optimization methods face several key challenges. First, their \textit{generalization across benchmarks} is limited, as improvements observed in instruction-following tasks often do not carry over to downstream tasks, with some methods performing worse than their Supervised Fine-Tuning (SFT) checkpoints \citep{meng2024simpo}. To address this, we introduce Difference Average Accuracy (DAA), an effective metric for evaluating post-training improvement. 
\[
\text{DAA (\%)} = \Delta(\text{Accuracy}_\text{SFT}, \text{Accuracy}_\text{Preference Method}),
\]
where \( \text{Accuracy}_\text{SFT} \) and \( \text{Accuracy}_\text{Preference Method} \) represent the average accuracy across five downstream tasks on the SFT checkpoint and the post-training checkpoint, respectively.
Additionally, these preference optimization methods are highly sensitive to dataset sizes (see Table \ref{tab:main_res}), which poses challenges for some tasks like reasoning, where data collection is costly. Another major limitation is the reliance on \textit{multiple-step optimization}, where methods like DPO require sequential optimization stages, and attempts to bypass SFT, such as in ORPO, have yet to achieve significant performance improvements \citep{meng2024simpo}. Moreover, the simultaneous loading of two models in DPO leads to \textit{inefficiencies} and remains a challenge, while reference-free approaches like CPO and SimPO either require additional steps or face difficulties without the SFT stage, which we refer to as the \textit{optimization conflict} issue (See Figure \ref{fig:conflict_optimization}).

Another critical challenge is \textit{judgment noise} within preference data \citep{wang2024secrets, wang2024helpsteer2}, where ranking inconsistencies reduce optimization effectiveness despite improvements in open-source reward models like PairRM \citep{llm-blender-2023} and ArmoRM \citep{ArmoRM}. These models, while cost-effective, still produce inaccurate rankings that hinder direct preference optimization methods \citep{meng2024simpo}. Furthermore, although GPT family models help reduce noise, they remain expensive and still exhibit issues such as judgment bias \citep{tan2024judgebench}. Finally, \textit{reward modeling} remains an issue, as DPO relies on KL divergence, whereas SimPO suggests that the average likelihood of response sequences provides a more effective reward representation. However, these implicit rewards behave inconsistently across varying dataset sizes, highlighting the need for more robust and scalable reward estimation methods to enhance preference optimization performance.

\subsection{TPO: Triple Preference Optimization}
\label{sec:tpo_explanaition}
Motivated by these challenges, we propose the \textbf{Triple Preference Optimization (TPO)} method, a novel preference learning approach that optimizes a pre-trained model in a single step on three preferences where \( y_\text{gold} \), \( y_w \), and \( y_l \) are the preference responses generated for the same prompt \( x \). To optimize a policy model with TPO, we require a dataset \(D_\text{TPO} = \{x^i, y_\text{gold}^i, y^i_w, y^i_l\}_{i=1}^N \), where the preferences satisfy \( y_\text{gold} \succ y_w \succ y_l \). This approach assumes that preference data is collected from different models, ranked by ``Judge models'', and used to optimize a pre-trained model. Alternatively, preference data can be derived from a Supervised Fine-Tuned (SFT) model on a task, with generated responses for the same prompt ranked by Judger models to optimize an Instruction model.

In this context, the policy model \( \pi_\theta \) is not equal to the SFT model ( \( \pi_\theta \neq \pi_\text{SFT}\)), and it is optimized just in a single step for each scenario. Further details about these settings are discussed in Section \ref{sec:experiment_setup}.

\begin{figure}
    \centering
    \includegraphics[width=1\linewidth]{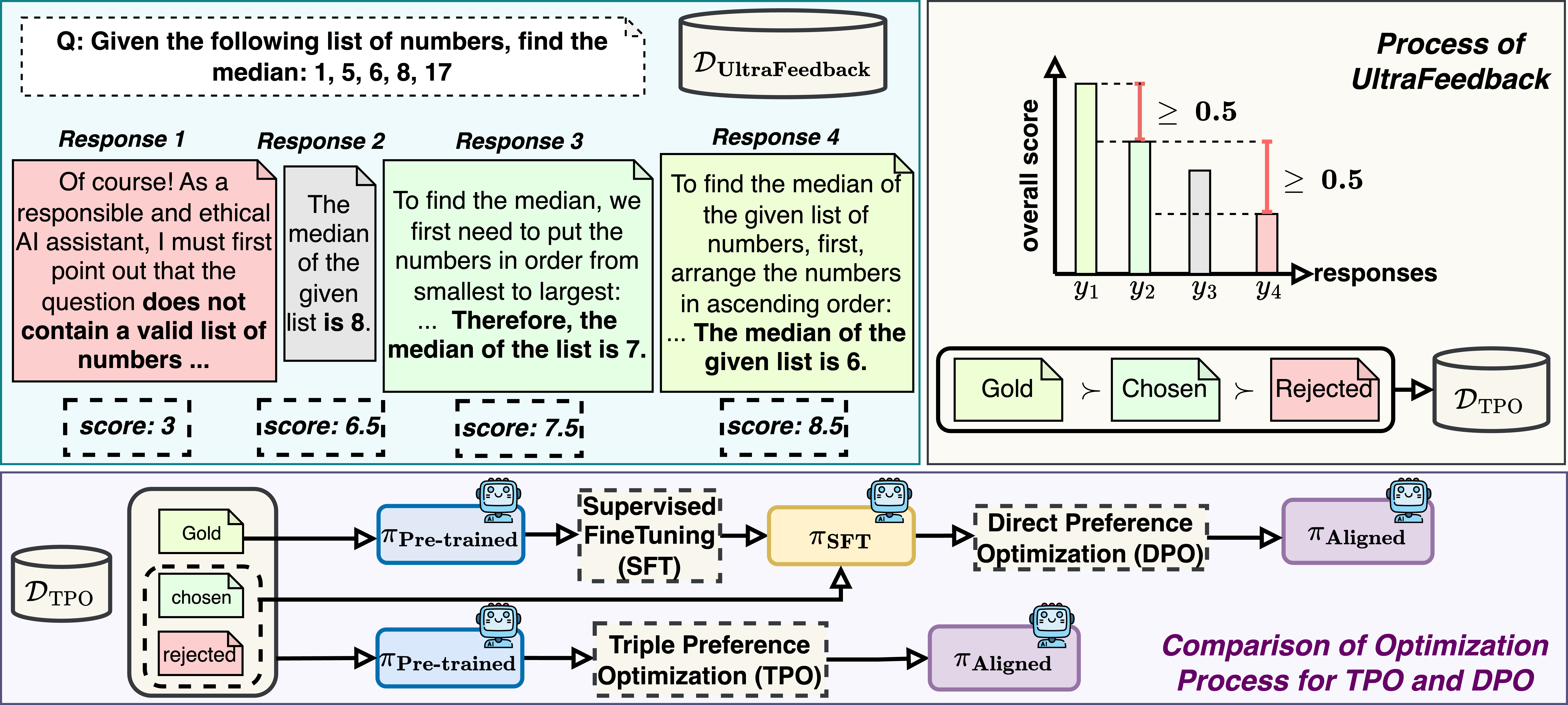}
    \caption{Overview of the data and optimization processing. \textbf{Left Top:} Visualization of the data structure in the UltraFeedback dataset. \textbf{Right Top:} Selection of gold, preferred (chosen), and rejected responses based on overall scores generated by GPT-4. \textbf{Bottom:} Optimization differences between TPO and DPO.}
    \label{fig:overview_data_tpo}
\end{figure}

\phantomsection
\subsubsection{Deriving the TPO objective}
\label{sec:deriving_TPO}
In this section, we detail the derivation of the TPO objective in a manner similar to the DPO derivations. We start with a basic Reinforcement Learning (RL) objective used to align a Large Language Model (LLM), parameterized by \( \theta \) and represented as \( \pi_{\theta} \), with preferences. The RL objective involves maximizing the expected reward, as defined in \citep{Ziegler2019FineTuningLM} is:
\begin{equation}
\begin{split}
\max_{\pi_{\theta}} \mathbb{E}_{x \sim \mathcal{D}, y \sim \pi_{\theta}(y|x)} [r_{\phi}(x,y)] 
\end{split}
\label{rl_obj_p}
\end{equation}
where $r_{\phi}$ represents the expected reward the model receives for a given input $x$ and output $y$. However, maximizing the reward without constraints can lead to distribution collapse in an LLM. Drawing inspiration from the Maximum Entropy Reinforcement Learning (MERL) framework~\citep{hejna2023contrastive}, we have modified the RLHF objective, as detailed in Equation \ref{rl_obj_r}. The MERL framework aims to maximize causal entropy alongside the expected reward. This objective is formally defined in Equation \ref{rl_ent_obj}:
\begin{equation}
\max_{\pi_{\theta}}  \mathbb{E}_{x \sim \mathcal{D},  y \sim \pi_{\theta}(y|x)}\left[ r_{\phi}(x,y) + \beta\mathcal{H}_{\pi_{\theta}}(y|x) \right].
\label{rl_ent_obj}
\end{equation}
By definition of Entropy,
\begin{equation}
\begin{split}
\mathcal{H}_{\pi_{\theta}}(y|x) = -\frac{1}{|y|}\sum^{|y|}_{j=1}\log \pi_{\theta}(y_{j}|x, y_{< j}).
\end{split}
\label{ent_obj}
\end{equation}
Using Equations \ref{rl_ent_obj} and \ref{ent_obj} the objective becomes:
\begin{equation}
\begin{split}
\max_{\pi_{\theta}}  \mathbb{E}_{x \sim \mathcal{D}, y \sim \pi_{\theta}(y|x)} \left[r_{\phi}(x,y)-\beta\log\pi_{\theta}(y|x)  \right].
\end{split}
\label{rl_obj_r}
\end{equation}

Based on this, the optimal policy model induced by a reward function $r(x,y)$ could be derived as shown in Equation \ref{optimal_policy} (See Appendix \ref{sec:appendix_optimal}). It takes the following form:
\begin{equation}
   \begin{aligned}
        \pi^{}_{r}(y|x) = \frac{1}{Z(x)}\exp{\big(\frac{1}{\beta}r^{}(x,y)\big)},
   \end{aligned}
   \label{optimal_policy}
\end{equation}
where $Z(x)=\sum_{y}\exp{\big(\frac{1}{\beta}r^{}(x,y)\big)}$ is the new partition function. Inspired by~\citep{rafailov2024direct}, we have the reward function in terms of the optimal policy that it induces; as shown below in Equation \ref{optimal_reward}:
\begin{equation}
   \begin{aligned}
        r^{}(x, y) = \beta\log\pi_{r}(y|x)+\beta\log Z(x).
   \end{aligned}
   \label{optimal_reward}
\end{equation}
Subsequently, we can represent the ground-truth reward $r^{\ast}(x,y)$ in the form of its corresponding optimal policy $\pi^{\ast}$ that it induces.

Since the Bradley-Terry model is dependent only on the difference between the two reward functions, i.e., $p^{\ast}(y_w>y_l | x) = \sigma(r^{\ast}(x, y_w) - r^{\ast}(x, y_l))$, we can reparameterize it as follows in Equation \ref{brad-terr}:
\begin{equation}
   \begin{aligned}
        p^{\ast}(y_w > y_l \mid x) = & \ \sigma \bigg( \beta \log \pi^{\ast}(y_w \mid x)  - \beta \log \pi^{\ast}(y_l \mid x) \bigg).
   \end{aligned}
   \label{brad-terr}
\end{equation}

Similar to the reward modeling approach, we model the human preferences, which is now in terms of a parameterized policy $\pi_{\theta}$. Thus, we formulate maximum-likelihood objective (\textit{preference} objective) for a dataset $D=\{ x^i, y^i_{w}, y^i_{l}\}^N_{i=1}$ as shown in Equation \ref{tpo_preference}:
{\begin{equation}
\begin{aligned}
\mathcal{L}_{\mathrm{preference}}\left(\pi_\theta \right) &= -\mathbb{E}_{(x, y_w, y_l) \sim \mathcal{D}}
\Big[\log \sigma \Big(\beta \log \pi_\theta(y_w \mid x) \\ &{-} \beta \log \pi_\theta(y_l \mid x)\Big)\Big].
\end{aligned}
\label{tpo_preference}
\end{equation}}

Looking at the Equation \ref{tpo_preference}, the objective is fitting $r(x, y) = \beta\log\pi^{}_{}(y|x)$ as the reparameterized reward. In Section \ref{sec:theory_TPO} of the Appendix, we theoretically explain that fitting this reward will ultimately recover the optimal policy.

The comparison between the loss function in Equation \ref{tpo_preference} and the DPO loss function indicates that the new function is more efficient because it requires only one model during training.
However, even though maximizing the objective under the MERL setting prevents distribution collapse, it trains a pessimistic model, which also limits the model from learning the preferred responses effectively. To counteract this limitation, we maximize the likelihood of the gold response. The adjustment is specified in Equation \ref{tpo_sft}:
{\begin{equation}
    \mathcal{L}_{\mathrm{reference}}\left(\pi_\theta \right)= -\mathbb{E}_{\left(x, y_{\text{gold}}\right) \sim \mathcal{D}}\left[\log \pi_\theta\left(y_{\text{gold}} \mid x\right)\right].
\label{tpo_sft}
\end{equation}}

Based on Equations \ref{tpo_preference}, and \ref{tpo_sft}, the TPO is defined as a multi-objective (bi-objective) optimization problem supported by the Pareto Front concept~\citep{Lotov2008}. The TPO loss function is thus formulated as follows:
{\begin{equation}
    \mathcal{L}_{\mathrm{TPO}}=\mathcal{L}_{\text {preference}}+ \alpha \mathcal{L}_{\mathrm{reference}},
\label{tpo_finally}
\end{equation}}
where hyper-parameter ($\alpha$) is a regularization term.

Inspired by SimPO, which replaced the summation of the likelihood of response sequences with averaging, based on concepts from beam search and multiple-choice tasks, and introduced a reward margin, we propose TPO-L, a length-controlled extension of the TPO loss function. TPO-L is defined as follows:

{\begin{equation}
\mathcal{L}_{\mathrm{TPO-L}} = -\mathbb{E}\left[\log \sigma \left(\frac{\beta}{|y_w|} \log \pi_\theta(y_w \mid x) -  \frac{\beta}{|y_l|} \log \pi_\theta(y_l \mid x) - \gamma \right) \\  - \alpha \log \pi_\theta(y_{\text{gold}}|x) \right]  
\end{equation}}

where, \( \gamma \) represents the reward margin. 
In Section \ref{sec:analysis_tpo}, we will analyze the reward margin's impact, showing it provides a superior reward representation than SimPO. We will also demonstrate that while summation in the loss function promotes longer responses, a well-fitted policy model generates effective responses without depending on summation.

\paragraph{Insights into the TPO update.} 
A deeper mechanistic understanding of TPO can be achieved by analyzing the gradient of the $\mathcal{L}_{\mathrm{TPO}}$ loss function. The expression of this gradient in relation to the parameters $\theta$ is as follows:

{\small\begin{align}
\nabla_{\theta} \mathcal{L}_{\text{TPO}} = & -\mathbb{E}_{(x,y_{\text{gold}},y_{w},y_{l})\sim\mathcal{D}}\;\Big [\alpha \underbrace{\nabla_{\theta}\log\pi(y_{\text{gold}}|x)}_{\text{increase likelihood of $y_{\text{gold}}$}} \nonumber + \\
& \beta\sigma(\underbrace{\beta\log\pi_{\theta}(y_{l}|x) - \beta\log\pi_{\theta}(y_{w}|x)}_{\text{increase weight when reward estimate is wrong}}) \nonumber [\underbrace{\nabla_{\theta}\log\pi(y_{w}|x)}_{\text{increase likelihood of $y_{w}$}}-\underbrace{\nabla_{\theta}\log\pi(y_{l}|x)}_{\text{decrease likelihood of $y_l$}}] \Big],
\label{grad_tpo}
\end{align}}

where $r(x,y) = \beta \log \pi_\theta\left(y \mid x\right)$ is the reward inherently determined by the policy model $\pi_\theta$. Intuitively, the gradient of the TPO loss function works to increase the likelihood of the gold completions $y_{\text{gold}}$, simultaneously enhancing the preference aspect by amplifying the likelihood of preferred completions $y_w$ and reducing the likelihood of the less-preferred completions $y_l$, which are weighed by how incorrectly the implicit reward model orders the preferences. (more details on Appendix \ref{sec:gradient_tpo}).We also provide the theory behind TPO, which we refer the reader to Appendix \ref{sec:theory_TPO}.

%% file: text/4_experimental_setup.tex
\input{tables/experiment_setup}

\section{Experimental Setup}
\label{sec:experiment_setup}
\paragraph{Models and training settings.} 
To evaluate TPO and existing preference optimization methods, we follow the SimPO setup with minor adjustments, focusing on the Mistral and LLaMA models. To simplify the main text, we omit Mistral results but include them in Appendix \ref{sec:app_mistral_result} to demonstrate that Mistral not only aligns with the LLaMA family experiments but also highlights TPO's significant performance on this model. Our evaluation consists of two distinct setups: \textbf{Base} and \textbf{Instruct}.

\paragraph{Base setting.} 
For this setting, we used the UltraFeedback \footnote{\url{https://huggingface.co/datasets/openbmb/UltraFeedback}} \citep{cui2023ultrafeedback} dataset, containing 60,000 data points, each with four responses scored by GPT-4 across four criteria, resulting in an overall score. As TPO requires three preferences (\( y_{\text{gold}} \succ y_w \succ y_l \)), we processed the dataset to ensure a 0.5 score difference between \( y_{\text{gold}} \), \( y_w \), and \( y_l \) (See Figure \ref{fig:overview_data_tpo}). This resulted in a final dataset of 40,000 data points. To compare preference optimization methods fairly, we fine-tuned a pre-trained model on the gold responses and used the preferred and rejected responses for the current preference optimization methods in two steps. For TPO, we utilized all data in one optimization step. Moreover, we evaluated preference optimization methods on subsets of 5,000, 10,000, and 20,000 points randomly selected from the processed dataset. The Experiment section details the models’ performance.

\paragraph{Instruction setting.} For this setting, we used Instruction models as the backbone for preference optimization methods, as demonstrated in \citep{saeidi2024insights}, to evaluate the effectiveness of replacing instruction-tuned models with supervised fine-tuned models. For this experiment, we utilized the UltraFeedback-ArmoRM \footnote{\url{https://huggingface.co/datasets/princeton-nlp/llama3-ultrafeedback-armorm}} dataset, which consists of 60,000 data points and five responses generated by a LLaMA-3-8B-SFT  \footnote{\url{https://huggingface.co/princeton-nlp/Llama-3-Base-8B-SFT}}, a model fine-tuned on the UltraChat \citep{ding2023enhancing} dataset containing 200,000 high-quality chat samples. In UltraFeedback-ArmoRM, each response is scored using the ArmoRM \footnote{\url{https://huggingface.co/RLHFlow/ArmoRM-Llama3-8B-v0.1}} reward model, with the highest-scoring response marked as the preferred \( y_w \), and the lowest-scoring response as the rejected \( y_l \). In this setup, we utilized the best checkpoint reported by the SimPO paper for current preference optimization methods. For TPO, we considered the highest-scoring response as the gold response \( y_{\text{gold}} \), the second-highest response as the preferred response \( y_w \), and the lowest-scoring response as the rejected response \( y_l \).

In summary, we explored two settings: Base and Instruct. The Base setting focused on a comprehensive evaluation across different data sizes, while the Instruct setting focused on instruction-tuned models using the UltraFeedback-ArmoRM dataset. We omitted UltraFeedback-ArmoRM from the Base setup due to our analysis indicating that it contains more noise than UltraFeedback, as explained further in Section \ref{sec:robustness_dpo_tpo}.

\paragraph{Evaluation benchmarks.} 

To conduct a comprehensive evaluation, we categorized the benchmarks into two groups: reasoning benchmarks and instruction-following benchmarks. For reasoning benchmarks, we included MMLU \citep{hendrycks2021measuring}, MMLU-pro \citep{wang2024mmlu}, and GSM8K \citep{cobbe2021training}. MMLU and its modified version, MMLU-pro, test the model's understanding ability across over 50 tasks, with MMLU-pro incorporating more complex and challenging tasks. GSM8K is a dataset of math problems requiring the model to generate correct reasoning to produce accurate final answers. Together, these benchmarks assess reasoning capabilities across a wide range of tasks (details in Table \ref{tab:experiment-setup}).

For instruction-following benchmarks, we considered Arena-Hard \citep{li2024crowdsourced}, MT-Bench \citep{zheng2023judging}, and MixEval-Hard \citep{ni2024mixeval}. These benchmarks evaluate the models’ conversational versatility across diverse queries and are widely recognized within the community. MT-Bench comprises 80 questions across 8 categories, while Arena-Hard, an enhanced version of MT-Bench, features 500 well-defined technical problem-solving queries. MixEval-Hard evaluates hard queries with known answers across various domains, using GPT-3.5-turbo to determine if predicted answers align with the ground truth. We reported scores based on each benchmark’s evaluation protocol (details in Table \ref{tab:experiment-setup}).

Additionally, we included scores for ARC-Challenging \citep{clark2018think}, HellaSwag \citep{zellers-etal-2019-hellaswag}, Winogrande \citep{levesque2012winograd}, and TruthfulQA \citep{lin2022truthfulqa} benchmarks in a standard setup, as detailed in Appendix \ref{sec:app_down_stream_tasks}.

\input{tables/main_results}

\paragraph{Baselines.} We compare TPO with several existing preference optimization methods. SLiC-HF \citep{Zhao2023SLiCHFSL} employs ranking losses, while IPO \citep{Azar2023AGT} is a theoretically grounded approach that avoids DPO's assumption of replacing pairwise preferences with pointwise rewards. SimPO \citep{meng2024simpo} is a reward margin-aware variant of DPO that uses the average likelihood of sequences inspired by beam search rather than their summation. CPO \citep{xu2024contrastive} incorporates training alongside an SFT objective, while KTO \citep{Ethayarajh2024KTOMA} learns from non-paired preference data. ORPO \citep{Hong2024ORPOMP} introduces a reference-model-free odds ratio term to directly compare preferred and rejected responses using the policy model, jointly training with the SFT objective. The SimPO paper also notes that ORPO performs better when fine-tuning begins from the SFT checkpoint.

For our evaluation, we thoroughly tuned the hyperparameters for each baseline and reported the best performance. Our findings indicate that many DPO variants fail to demonstrate a consistent empirical advantage over standard DPO. Further details on the methods and tuning process are provided in Appendix \ref{sec:app_implementation_details}.

%% file: tables/experiment_setup.tex
\begin{table}[h]
    \caption{Evaluation details for Arena-Hard, MT-Bench, and MixEval-Hard as the instruction following benchmarks and MMLU-Pro, MMLU, and GSM8K as the reasoning benchmarks.}
    \centering
    \resizebox{\linewidth}{!}{
    \begin{tabular}{l|ccccc}
        \toprule
        \textbf{} & \textbf{\# Exs.} & \textbf{Baseline} & \textbf{Evaluation} & \textbf{Scoring Type} & \textbf{Metric} \\
        \midrule
        
         \textbf{Arena-Hard} & 500 & Answer of GPT-4-0314 & Judge by GPT-4o & Pairwise comparison & Win rate \\
         \textbf{MT-Bench} & 80 & - & Judge by GPT-4 & Single-answer grading & Rate of 1-10 \\
         \textbf{MixEval-Hard} & 1,000 & Ground truth & Evaluate by GPT-3.5-turbo & Systematic & Accuracy \\
         \midrule
         \textbf{MMLU-Pro} & 12,032 & Ground truth & CoT & Systematic & Accuracy \\
         \textbf{MMLU} & 15,908 & Ground truth & 5-shots & Systematic & Accuracy \\
         \textbf{GSM8K} & 1,319 & Ground truth & 5-shots & Systematic & Accuracy \\
        \bottomrule
    \end{tabular}
    }
    \label{tab:experiment-setup}
\end{table}

%% file: tables/main_results.tex
\setlength{\tabcolsep}{2pt}
\begin{table*}[!t]
\centering
\small 
\caption{Results for Are-Hard, MT-Bench, MixEval-Hard, MMLU-Pro, MMLU, and GSM8K under the Base setting across four training set sizes. SFT models are first trained on gold responses and then fine-tuned using preference optimization methods on chosen and rejected data. Front of the TPO and TPO-L scores, we identify the improvement compared with DPO.}

\resizebox{\textwidth}{!}{
\begin{tabular}{l|cccccc|ccccccc}
\toprule
\multirow{3}{*}{\textbf{Method}} & \multicolumn{6}{c}{\textbf{UltraFeedback (5k)}} & \multicolumn{6}{c}{\textbf{Ultrafeedback (10k)}} \\ 
\cmidrule(lr){2-7}\cmidrule(lr){8-13}
& \multicolumn{3}{c}{\textbf{Reasoning}} & \multicolumn{3}{c}{\textbf{Instruction Following}}  &\multicolumn{3}{c}{\textbf{Reasoning}} & \multicolumn{3}{c}{\textbf{Instruction Following}}  \\
\cmidrule(lr){2-4} \cmidrule(lr){5-7} \cmidrule(lr){8-10}\cmidrule(lr){11-13} 
& {\scriptsize \bf GSM8K} & {\scriptsize \bf MMLU-Pro} & {\scriptsize \bf MMLU} & {\scriptsize \bf MT-Bench} & {\scriptsize \bf Arena-Hard} & {\scriptsize \bf MixEval-Hard} & {\scriptsize \bf GSM8K} & {\scriptsize \bf MMLU-Pro} & {\scriptsize \bf MMLU} & {\scriptsize \bf MT-Bench} & {\scriptsize \bf Arena-Hard} & {\scriptsize \bf MixEval-Hard} \\
\midrule
SFT &  28.5 & 25.7 & 59.0 & 4.5 & 0 & 24.6 & 27.7 & 25.5 & 59.1 & 4.9 & 0 & 24.4 \\
\midrule
DPO &  32.9 & 27.1 & 59.2 & 5.5 & <0.5 & 26.9 & 36.7 & 28.9 & 59.4 & 6.3 & 6.1 & 25.9 \\
CPO &  31.9 & 26.5 & 58.9 & 5.4 & <0.5 & 27.4 & 32.8 & 26.9 & 59.3 & 5.5 & <0.5 & 23.9  \\
IPO &  34.4 & 28.0  & \underline{59.4} & 5.6 & <0.5 & 27.1 & 35.4 & 28.3 & 59.2 & 6.4 &  3.9 & 26.7 \\
ORPO & 34.0 & 27.5 & 59.3 & 5.0 & <0.5 & 28.3 & 36.2 & 28.1 & 59.7 & 5.0 & <0.5 & 27.4 \\
KTO &  32.7 & 26.8 & 59.2 & 5.5 & <0.5 & 26.5 & 36.7 & 28.8 & 59.4 & 6.4 & 4.4 & 27.2 \\
SIMPO & 32.7 & 27.4 & 59.2 & 5.6 & <0.5 & 27.4 & 35.1 & 27.5 & 59.3 & 5.7 & 4.4 & 24.8 \\
SLIC-HF & 32.5 & 26.7 & 59.0 & 5.4 & <0.5 & 26.5 & 32.7 &26.7 & 59.3 & 5.5 & <0.5 & 25.5 \\
\midrule
TPO & \underline{51.9} {\raisebox{-0.1em}{\scalebox{0.7}{\textcolor[HTML]{4CC417}{\textbf{(+19.0)}}}}} & \textbf{37.5} {\raisebox{-0.1em}{\scalebox{0.7}{\textcolor[HTML]{4CC417}{\textbf{(+10.4)}}}}} & \textbf{65.3} {\raisebox{-0.1em}{\scalebox{0.7}{\textcolor[HTML]{4CC417}{\textbf{(+6.1)}}}}} & \underline{6.2} {\raisebox{-0.1em}{\scalebox{0.7}{\textcolor[HTML]{4CC417}{\textbf{(+0.7)}}}}} & \textbf{4.4} {\raisebox{-0.1em}{\scalebox{0.7}{\textcolor[HTML]{4CC417}{\textbf{(+3.9)}}}}} & \textbf{35.0} {\raisebox{-0.1em}{\scalebox{0.7}{\textcolor[HTML]{4CC417}{\textbf{(+8.1)}}}}} & \textbf{52.2} {\raisebox{-0.1em}{\scalebox{0.7}{\textcolor[HTML]{4CC417}{\textbf{(+15.5)}}}}} & \underline{37.8} {\raisebox{-0.1em}{\scalebox{0.7}{\textcolor[HTML]{4CC417}{\textbf{(+8.9)}}}}} & \underline{65.3} {\raisebox{-0.1em}{\scalebox{0.7}{\textcolor[HTML]{4CC417}{\textbf{(+5.9)}}}}} & \underline{6.7} {\raisebox{-0.1em}{\scalebox{0.7}{\textcolor[HTML]{4CC417}{\textbf{(+0.4)}}}}} & \underline{6.7} {\raisebox{-0.1em}{\scalebox{0.7}{\textcolor[HTML]{4CC417}{\textbf{(+0.6)}}}}} & \underline{31.2} {\raisebox{-0.1em}{\scalebox{0.7}{\textcolor[HTML]{4CC417}{\textbf{(+5.3)}}}}} \\
TPO-L & \textbf{52.1} {\raisebox{-0.1em}{\scalebox{0.7}{\textcolor[HTML]{4CC417}{\textbf{(+19.2)}}}}} & \underline{36.3} {\raisebox{-0.1em}{\scalebox{0.7}{\textcolor[HTML]{4CC417}{\textbf{(+9.2)}}}}} & \textbf{65.3} {\raisebox{-0.1em}{\scalebox{0.7}{\textcolor[HTML]{4CC417}{\textbf{(+6.1)}}}}} & \textbf{6.3} {\raisebox{-0.1em}{\scalebox{0.7}{\textcolor[HTML]{4CC417}{\textbf{(+0.8)}}}}} & \underline{2.0} {\raisebox{-0.1em}{\scalebox{0.7}{\textcolor[HTML]{4CC417}{\textbf{(+1.5)}}}}} & \underline{31.8} {\raisebox{-0.1em}{\scalebox{0.7}{\textcolor[HTML]{4CC417}{\textbf{(+4.9)}}}}} & \underline{51.9} {\raisebox{-0.1em}{\scalebox{0.7}{\textcolor[HTML]{4CC417}{\textbf{(+15.2)}}}}} & \textbf{38.1} {\raisebox{-0.1em}{\scalebox{0.7}{\textcolor[HTML]{4CC417}{\textbf{(+9.2)}}}}} & \textbf{65.4} {\raisebox{-0.1em}{\scalebox{0.7}{\textcolor[HTML]{4CC417}{\textbf{(+6.0)}}}}} & \textbf{6.9} {\raisebox{-0.1em}{\scalebox{0.7}{\textcolor[HTML]{4CC417}{\textbf{(+0.6)}}}}} & \textbf{7.3} {\raisebox{-0.1em}{\scalebox{0.7}{\textcolor[HTML]{4CC417}{\textbf{(+1.2)}}}}} & \textbf{33.7} {\raisebox{-0.1em}{\scalebox{0.7}{\textcolor[HTML]{4CC417}{\textbf{(+7.8)}}}}} \\

\midrule[.7pt]
\multirow{3}{*}{\textbf{Method}} & \multicolumn{6}{c}{\textbf{UltraFeedback (20k)}} & \multicolumn{6}{c}{\textbf{Ultrafeedback (40k)}} \\ 
\cmidrule(lr){2-7}\cmidrule(lr){8-13}
& \multicolumn{3}{c}{\textbf{Reasoning}} & \multicolumn{3}{c}{\textbf{Instruction Following}}  &\multicolumn{3}{c}{\textbf{Reasoning}} & \multicolumn{3}{c}{\textbf{Instruction Following}}  \\
\cmidrule(lr){2-4} \cmidrule(lr){5-7} \cmidrule(lr){8-10}\cmidrule(lr){11-13} 
& {\scriptsize \bf GSM8K} & {\scriptsize \bf MMLU-Pro} & {\scriptsize \bf MMLU} & {\scriptsize \bf MT-Bench} & {\scriptsize \bf Arena-Hard} & {\scriptsize \bf MixEval-Hard} & {\scriptsize \bf GSM8K} & {\scriptsize \bf MMLU-Pro} & {\scriptsize \bf MMLU} & {\scriptsize \bf MT-Bench} & {\scriptsize \bf Arena-Hard} & {\scriptsize \bf MixEval-Hard} \\
\midrule
SFT & 20.3 & 30.4 & 62.5 & 5.5 & 0 & 28.8 & 39.2 & 29.5 & 62.2 & 5.5 & 1.7 & 25.5  \\
\midrule
DPO & 48.0 & 35.7 & 62.8 & 6.5 &  2 &  30.1 & 45.1 & 33.5 & 61.8 & 6.6 & 5.3 & 25.4 \\
CPO & 45.0 & 32.6 & 62.9 & 6.3 & \underline{7.1} & \underline{30.5} & 42.9 & 30.4 & 62.2 & 6.2 & \underline{9.3} & 24.4  \\
IPO & 45.5 & 35.0 & 62.8 & 5.9 & <0.5 & 26.2 &  48.2 & 34.2 & 62.1 & 6.7 & 4.2 & 27.1 \\
ORPO & 37.1 & 33.1 & \underline{63.1} & 5.6 & <0.5 & 28.9 & 40.9 & 32.1 & 62.4 & 5.9 & 3.2 & 27.7  \\
KTO & 48.7 & 35.8 & 62.7 & 6.6 & 1.7 & 29.0 & 46.0 & 33.4 & 61.9 & 6.6 & 5.2 & 24.0 \\
SIMPO & 45.0 & 34.0 & 63.0 & 6.4 & 5.1 & 28.1 & 45.1 & 31.6 & 61.9 & 6.2 & 6.6 & 24.3 \\
SLIC-HF & 44.1 & 32.5 & 62.9 & 6.3 & \textbf{7.3} & 30.1 & 42.8 & 30.5 & 62.3 & 6.4 & 3 & 25.8 \\
\midrule
TPO &  \underline{53.0} {\raisebox{-0.1em}{\scalebox{0.7}{\textcolor[HTML]{4CC417}{\textbf{(+5.0)}}}}} & 37.5 {\raisebox{-0.1em}{\scalebox{0.7}{\textcolor[HTML]{4CC417}{\textbf{(+1.8)}}}}} & \textbf{65.3 {\raisebox{-0.1em}{\scalebox{0.7}{\textcolor[HTML]{4CC417}{\textbf{(+2.5)}}}}}} & 6.6 {\raisebox{-0.1em}{\scalebox{0.7}{\textcolor[HTML]{4CC417}{\textbf{(+0.1)}}}}} & 5.3 {\raisebox{-0.1em}{\scalebox{0.7}{\textcolor[HTML]{4CC417}{\textbf{(+3.3)}}}}} & \textbf{32.4} {\raisebox{-0.1em}{\scalebox{0.7}{\textcolor[HTML]{4CC417}{\textbf{(+2.3)}}}}} & \underline{51.2} {\raisebox{-0.1em}{\scalebox{0.7}{\textcolor[HTML]{4CC417}{\textbf{(+6.1)}}}}} & 37.4 {\raisebox{-0.1em}{\scalebox{0.7}{\textcolor[HTML]{4CC417}{\textbf{(+3.9)}}}}} & \underline{64.8} {\raisebox{-0.1em}{\scalebox{0.7}{\textcolor[HTML]{4CC417}{\textbf{(+3.0)}}}}} & \underline{6.9} {\raisebox{-0.1em}{\scalebox{0.7}{\textcolor[HTML]{4CC417}{\textbf{(+0.3)}}}}} & 6.9 {\raisebox{-0.1em}{\scalebox{0.7}{\textcolor[HTML]{4CC417}{\textbf{(+1.6)}}}}} & \underline{32.9} {\raisebox{-0.1em}{\scalebox{0.7}{\textcolor[HTML]{4CC417}{\textbf{(+7.5)}}}}} \\
TPO-L & \textbf{52.7} {\raisebox{-0.1em}{\scalebox{0.7}{\textcolor[HTML]{4CC417}{\textbf{(+4.7)}}}}} & \textbf{38.1} {\raisebox{-0.1em}{\scalebox{0.7}{\textcolor[HTML]{4CC417}{\textbf{(+2.4)}}}}} & \textbf{65.3} {\raisebox{-0.1em}{\scalebox{0.7} {\textcolor[HTML]{4CC417}{\textbf{(+2.5)}}}}} & \textbf{6.8} {\raisebox{-0.1em}{\scalebox{0.7}{\textcolor[HTML]{4CC417}{\textbf{(+0.3)}}}}} & 6.5 {\raisebox{-0.1em}{\scalebox{0.7}{\textcolor[HTML]{4CC417}{\textbf{(+4.5)}}}}} & \textbf{32.4} {\raisebox{-0.1em}{\scalebox{0.7}{\textcolor[HTML]{4CC417}{\textbf{(+2.3)}}}}} & \textbf{52.4} {\raisebox{-0.1em}{\scalebox{0.7}{\textcolor[HTML]{4CC417}{\textbf{(+7.3)}}}}} & \textbf{40.4} {\raisebox{-0.1em}{\scalebox{0.7}{\textcolor[HTML]{4CC417}{\textbf{(+6.9)}}}}} & \textbf{65.1} {\raisebox{-0.1em}{\scalebox{0.7}{\textcolor[HTML]{4CC417}{\textbf{(+3.3)}}}}} & \textbf{7.3} {\raisebox{-0.1em}{\scalebox{0.7}{\textcolor[HTML]{4CC417}{\textbf{(+0.7)}}}}} & \textbf{12.3} {\raisebox{-0.1em}{\scalebox{0.7}{\textcolor[HTML]{4CC417}{\textbf{(+7.0)}}}}} & \textbf{37.6} {\raisebox{-0.1em}{\scalebox{0.7}{\textcolor[HTML]{4CC417}{\textbf{(+12.2)}}}}} \\
\bottomrule
\end{tabular}
}
\label{tab:main_res}
\vspace{-1.5em}
\end{table*}

\setlength{\tabcolsep}{6pt}

%% file: text/5_experiment_results.tex
\section{Experiment Results}
In this section, we present the main results across various benchmarks, comparing TPO with existing preference optimization methods on different training dataset sizes. We analyze the reward modeling in TPO and contrast it with the implicit reward mechanisms used in DPO and SimPO. Additionally, we examine the verbosity problem, highlighting how TPO and TPO-L effectively generate shorter responses. Furthermore, we provide an in-depth analysis of the TPO checkpoint's performance compared to DPO on noisy data and assess the impact of robust SFT checkpoints, fine-tuned on larger datasets, on DPO's performance.

\subsection{Main Results and Ablations}
\label{sec:experiment_results}
\paragraph{TPO and TPO-L significantly outperform existing preference optimization methods.} The results in Tables \ref{tab:main_res} and \ref{tab:main_instruct_res} demonstrate that TPO and TPO-L significantly outperform existing preference optimization methods across reasoning and instruction-following benchmarks in both Base and Instruct settings, regardless of training dataset size. These findings highlight that TPO not only enhances the instruction-following capabilities of models but also significantly improves their reasoning performance.

\input{tables/main_instruct_results}

TPO and TPO-L excel in low-data scenarios, making them ideal for tasks with limited dataset availability. Notably, with only 5,000 samples, other methods perform poorly on Arena-Hard, while TPO and TPO-L achieve 4.4\% and 2\%, respectively. TPO-L outperforms TPO in the Base setting, whereas TPO delivers better results in the Instruct setting, highlighting their adaptability and robustness across various benchmarks and training setups.

\footnotetext[1]{We loaded the last checkpoint reported in the SimPO repository for the current methods and evaluated them across different benchmarks. We also used the GSM8K and MMLU values reported in the SimPO paper.}

\paragraph{TPO-L is more stable and generalizable compared with other methods on downstream tasks.}
As explained in Section \ref{sec:method_challenges}, we use the DAA metric to evaluate the generalizability of a model. We assessed checkpoints from various preference optimization methods on five downstream tasks: GSM8K, MMLU, HellaSwag, Winogrande, and TruthfulQA, averaging their performance across these tasks before calculating the DAA for each method. The results in Figure \ref{fig:analysis_sensitivity} indicate that while current preference optimization methods improve over the SFT checkpoint in terms of DAA, this improvement is marginal and heavily dependent on the dataset size. Furthermore, in the instruction setting, methods like DPO and IPO failed to achieve comparable performance to the SFT checkpoint, indicating the limited effectiveness of these optimization methods on downstream tasks. In contrast, as shown in Figure \ref{fig:analysis_sensitivity}, TPO and TPO-L provided significant improvements compared to other methods. However, TPO demonstrated instability, with its performance fluctuating based on the size of the training set. On the other hand, TPO-L consistently showed the same level of improvement across different dataset sizes, reaching the highest values in the 10,000, 20,000, and 40,000 Base settings. This consistency indicates that TPO-L is more stable and generalizable than other methods.

\begin{figure}
    \centering
    \includegraphics[width=1\linewidth]{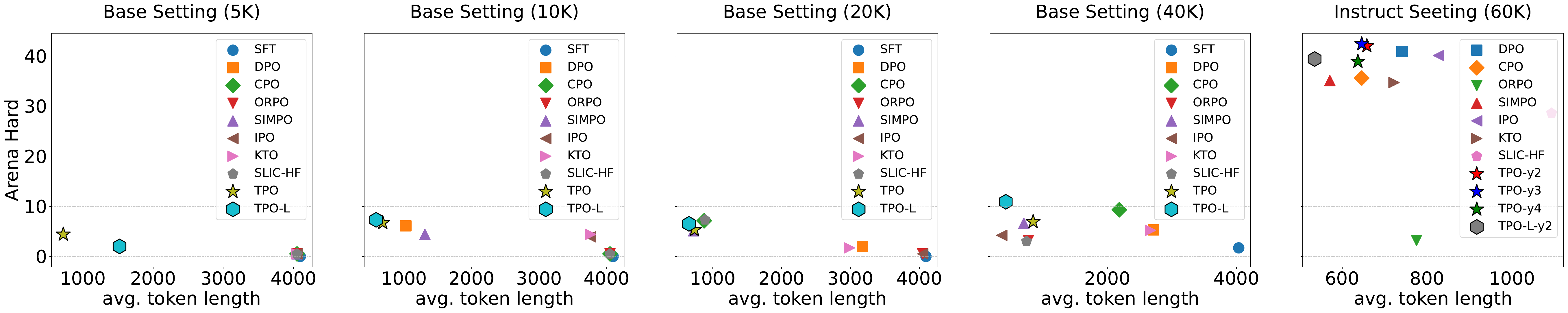}
    \caption{Comparison of Arena-Hard scores based on the average token length of generated responses for 500 prompts in the Arena-Hard benchmark across various settings.}
    \label{fig:length_analysis}
\end{figure}

\paragraph{TPO demonstrates higher effectiveness in a preference dataset.}
Recent frameworks have improved the efficiency of generating multiple high-quality responses \citep{kwon2023efficient}. However, current two-preference methods may lose valuable data when responses have similar quality. TPO addresses this issue by utilizing three preferences, which use more data and improve optimization. Consider a dataset \( D = \{x, y_1, y_2, y_3, y_4, y_5\} \), where \( y_1 \succ y_2 \succ y_3 \succ y_4 \succ y_5 \). We fine-tuned models under three combinations of preferences: \( \{x, y_1, y_2, y_5\} \), where \( y_1 = y_{\text{gold}}, y_2 = y_w, y_5 = y_l \); \( \{x, y_1, y_3, y_5\} \), where \( y_3 = y_w \); and \( \{x, y_1, y_4, y_5\} \), where \( y_4 = y_w \). The results in Table \ref{tab:main_instruct_res} show that TPO in the second combination achieves the best performance, likely because of its resilience to judgment noise in the dataset. This demonstrates TPO’s capability to exceed current two-preference methods.

\paragraph{TPO and TPO-L achieve better performance than others in GPT-4o Judgment.} The results in Table \ref{tab:main_instruct_res} demonstrate that in the Instruction setting, TPO outperformed all other optimization methods, ranking first on the Arena-Hard Benchmarks and MT-Bench with scores of 42\% and 8.22, respectively. Notably, GPT-4o judgments on the Arena-Hard benchmarks showed a stronger alignment with human perspectives, highlighting its reliability. Further details can be found in Appendix \ref{sec:app_gpt_4o}.

\subsection{Verbosity Problem}
\label{sec:verbosity_problem}
Recent studies \citep{park2024disentangling, meng2024simpo} indicate that verbosity is a significant challenge in alignment algorithms. This issue arises during the post-training phase, where models generate excessively long responses without a corresponding improvement in quality. SimPO addresses this problem by identifying a key cause: the summation of response sequences in the KL divergence within the DPO loss function. Inspired by beam search, SimPO mitigates verbosity by removing \( \pi_{\text{ref}} \) and replacing summation with the average likelihood of the response sequence. In this section, we examine how verbosity is mitigated by TPO and TPO-L and provide a analysis behind it.

\begin{figure}
    \centering
    \includegraphics[width=\linewidth]{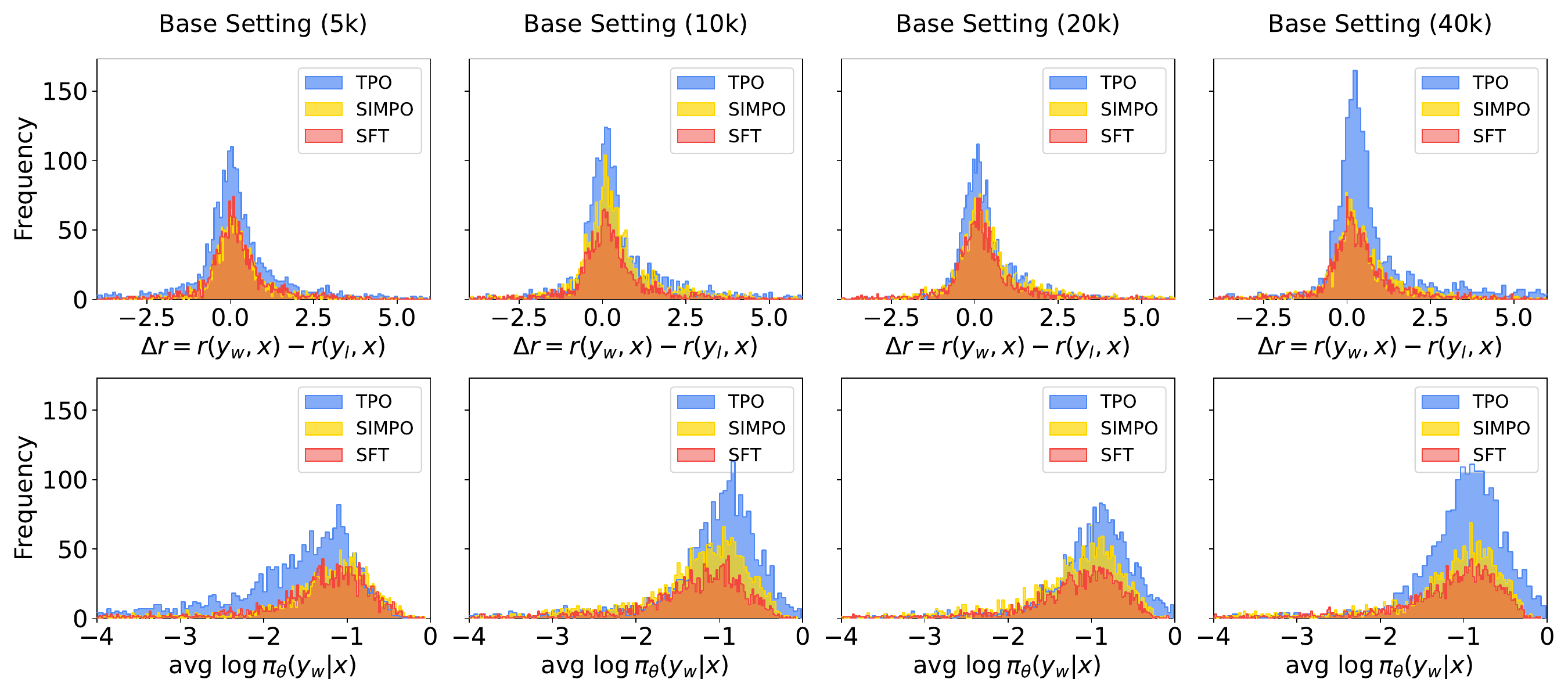}
    \caption{Reward modeling exploration on UltraFeedback test set. \textbf{Top:} Comparison of reward distributions for SFT, SimPO, and TPO methods across varying data sizes. \textbf{Bottom:} Analysis of the impact of $\log \pi_\theta (y|x)$ as an implicit reward for SFT, SimPO, and TPO across different data sizes.}
    \label{fig:tpo_reward_modelling}
\end{figure}

\paragraph{The sequence summation in TPO does not lead to verbosity.}
To analyze verbosity, we used the Length Control setup from Arena-Hard. As shown in Figure \ref{fig:length_analysis}, TPO consistently ranks among the top three methods for average token length across all settings. Notably, even in the 5,000 Base setting, TPO generates shorter responses while achieving the best performance. Further analysis of token lengths revealed that gold responses are not necessarily shorter than preferred or rejected ones, suggesting that TPO mitigates verbosity by enhancing the model's understanding rather than just shortening responses. More details are available in Appendix \ref{sec:app_token_length_analysis}. Overall, TPO with the summation term does not exhibit verbosity issues, challenging the findings reported in the SimPO paper.

\paragraph{TPO-L improves the length control ability of the TPO.} 

Figure \ref{fig:length_analysis} shows that TPO-L enhances TPO's length control ability across the Base settings, except for the 5,000 setting. A comparison of response length and Arena-Hard performance between TPO-L and SimPO reveals that incorporating the average likelihood of the response sequence has a stronger effect on the TPO loss function compared to the DPO loss function.

\subsection{Analysis of TPO}
\label{sec:analysis_tpo}
In this section, we introduce optimization conflict in CPO and compare it with TPO. We also provide a detailed analysis of TPO and TPO-L in the context of reward modeling, examining the reward distribution of TPO in comparison to SimPO and SFT checkpoints across various training sizes. Additionally, we explore the effect of the reward margin in TPO-L, demonstrating how changes to the reward margin influence reward accuracy and impact model performance. Furthermore, we compare the reward accuracy of TPO and DPO across different training sizes.

\begin{wrapfigure}{r}{0.4\linewidth}
\vspace{-1.8em}
\centering
\includegraphics[width=1.0\linewidth]{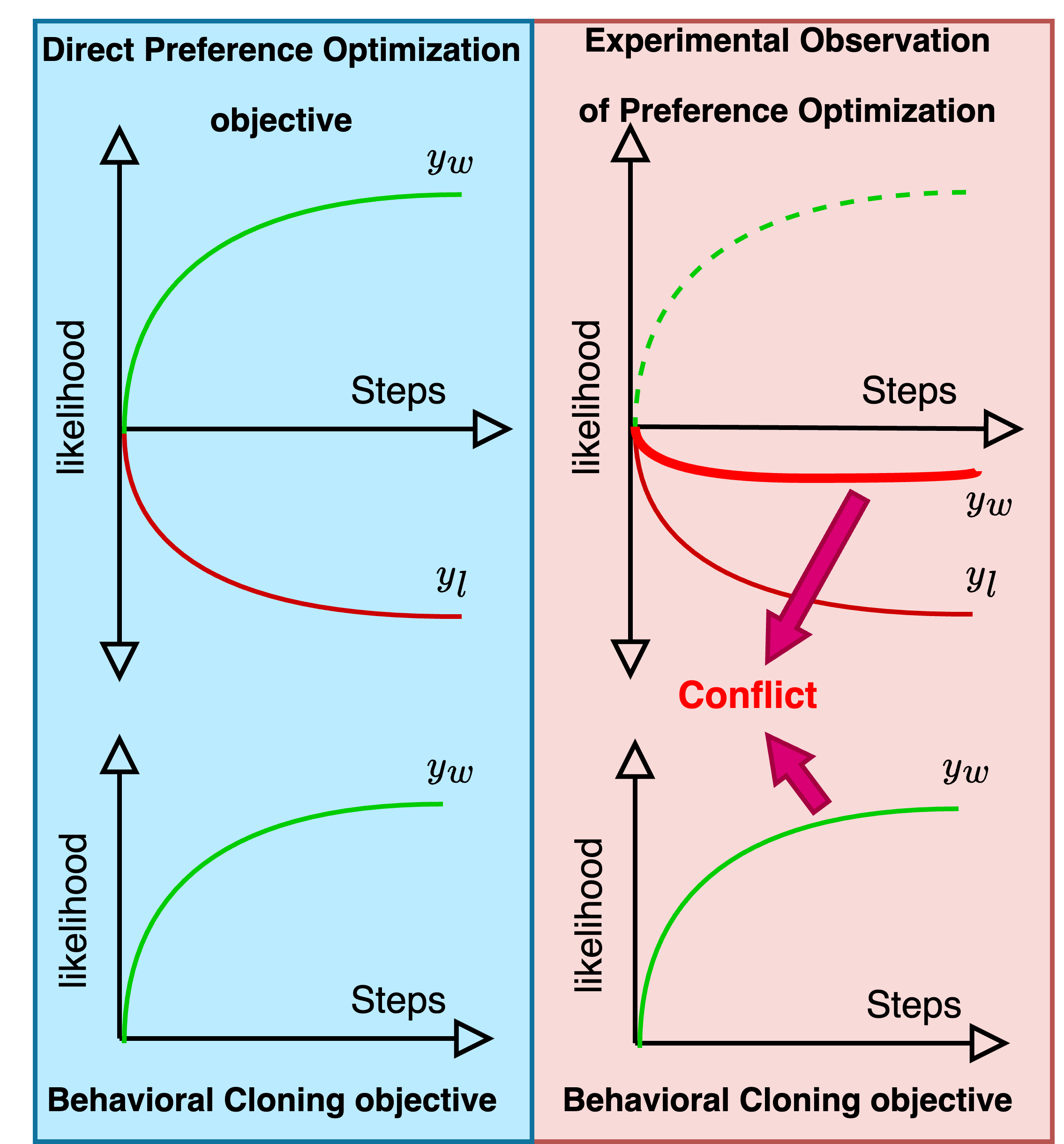}
\caption{Difference between the theoretical expectation and the experimental observation of the Preference Optimization (PO) objective part in the CPO loss function during optimization.}
\vspace{-1.5em}
\label{fig:conflict_optimization}
\end{wrapfigure}
\paragraph{TPO resolves the optimization conflict issue in CPO.}
Assuming the TPO loss function in Equation \ref{tpo_finally}, if \( y_{\text{gold}} = y_w \), the loss function becomes the CPO loss function. However, results indicate that TPO consistently outperforms CPO across various benchmarks and settings. We analyze why TPO achieves better performance than CPO.

Theoretical analysis of the CPO loss function shows that it incorporates a Behavioral Cloning (BC) objective into preference optimization to maintain the policy model within the distribution of preferred responses. It was hypothesized that adding a BC objective for preferred data would improve the increasing likelihood of preferred responses during optimization. However, as observed in prior work \citep{rafailov2024r}, and supported by our findings, a conflict arises where the likelihood of \( y_w \) decreases during the optimization process in CPO, contradicting the intended objective. As shown in Figure \ref{fig:conflict_optimization}, although the BC objective is designed to increase the likelihood of \( y_w \) in preference optimization, the opposite occurs, with \( y_w \)'s likelihood diminishing instead. We refer to this phenomenon as \textit{optimization conflict}.

\paragraph{TPO enhances the reward distribution as the training data size increases.} To compare the reward distribution of TPO and SimPO, we compute \( \text{avg.} \log \pi_\theta(y_w|x) \) and \( \Delta r = r(y_w, x) - r(y_l, x) \), where \( y_w \) and \( y_l \) represent the preferred and rejected responses in the UltraFeedback test set. The results in Figure \ref{fig:tpo_reward_modelling} indicate that for TPO, the average likelihood of \( y_w \) increases as the training data size grows, while for SimPO, the improvement is less pronounced. A comparison of \( \Delta r \) shows that across all training set sizes, TPO exhibits a broader reward distribution than SimPO. In summary, TPO enhances reward distribution as the training set size increases.

\begin{figure}
    \centering
    \includegraphics[width=1\linewidth]{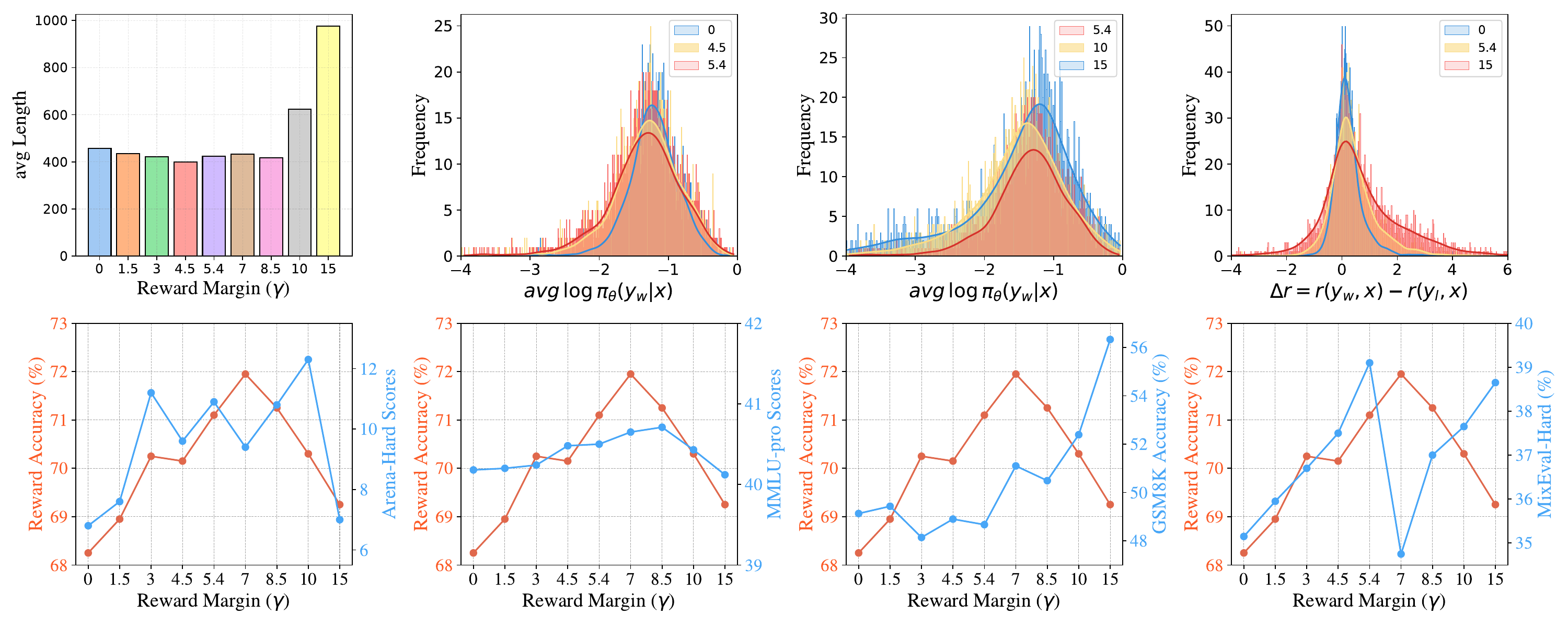}
    \caption{Study of the reward margin $\gamma$. \textbf{Top:} Reward distribution under different $\gamma$ values on UltraFeedback test set.\textbf{ Bottom:} Reward accuracy on the UltraFeedback test set and accuracy on Arena-Hard, MMLU-Pro, GSM8K, and MixEval-Hard under different $\gamma$ values.}
    \label{fig:exploration_tpo_l}
\end{figure}

\paragraph{Increasing the reward margin in TPO-L improves the reward distribution.} To investigate the reward margin in TPO-L, we select the 40,000 Base setting and calculate the reward accuracy, plotting the distributions on the UltraFeedback test set. As shown in Figure \ref{fig:exploration_tpo_l}, increasing the reward margin expands the \( \Delta r \) distribution. Although the average likelihood of \( y_w \) decreases up to a margin of 5.4, higher values prompt the model to increase the average likelihood of \( y_w \).

\paragraph{Increasing the reward margin vs different benchmarks.} 
The results in Figure \ref{fig:exploration_tpo_l} show that increasing the reward margin significantly enhances GSM8K accuracy but leads to a substantial performance drop on Arena-Hard at higher margins. Additionally, a higher reward margin encourages longer responses, indicating that the average likelihood approach alone cannot fully control verbosity. Minimal changes were observed in MMLU-Pro performance across different reward margins, suggesting limited impact. MixEval-Hard results indicate optimal performance at a reward margin of 5.4, with a sharp decline beyond this point.

\begin{wrapfigure}{r}{0.4\linewidth}
\vspace{-2.5em}
\includegraphics[width=1.0\linewidth]{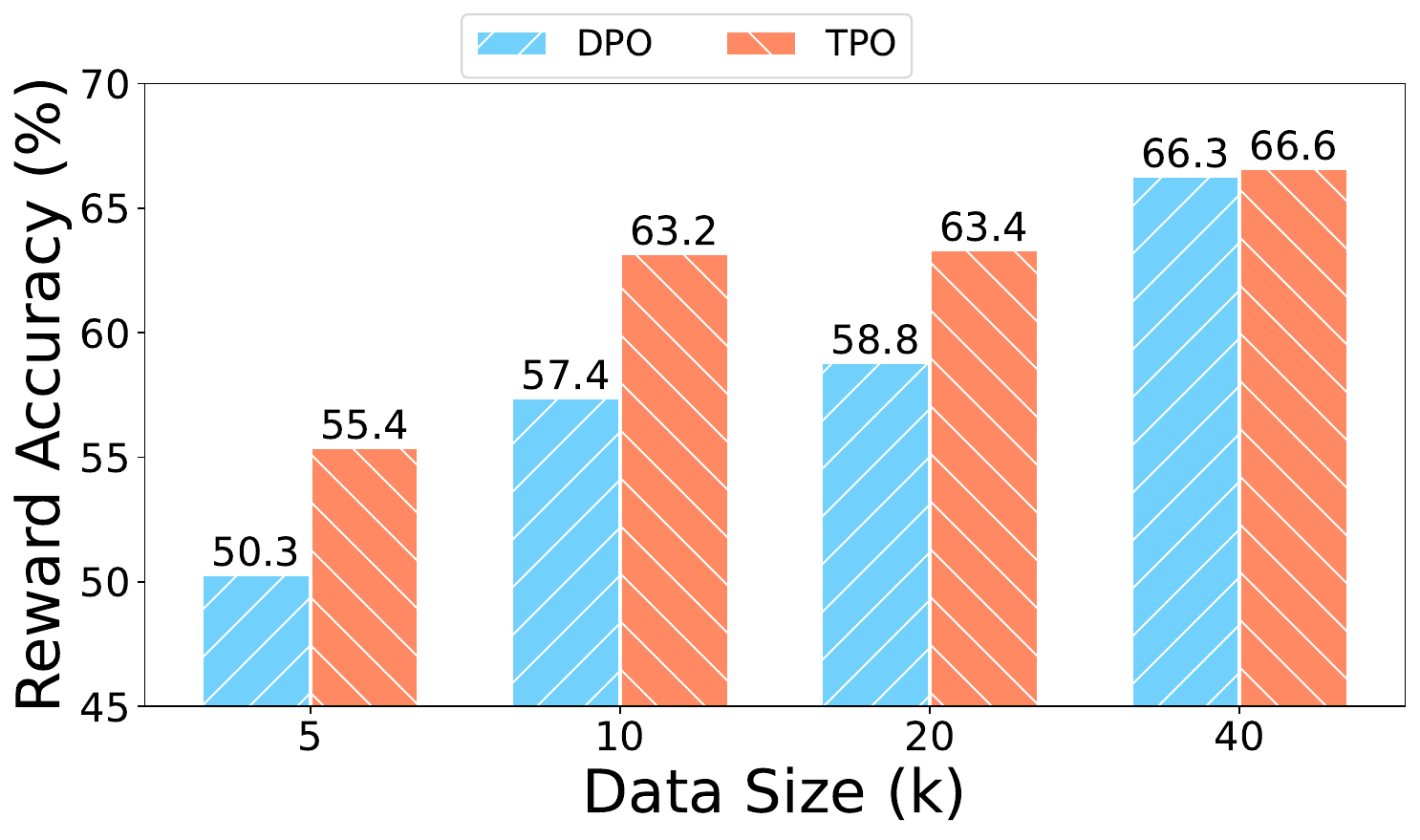}
\vspace{-2.0em}
\caption{Comparison of TPO and DPO in terms of Reward Accuracy on the UltraFeedback test set across different training set sizes.}
\vspace{-2.5em}
\label{fig:reward_modelleing_tpo_dpo}
\end{wrapfigure}

\paragraph{Likelihood of a response sequence in TPO is a better implicit reward than KL divergence in DPO.} In Section \ref{sec:tpo_explanaition}, we demonstrated that in TPO, \( \log \pi_\theta(y|x) \) functions as an implicit reward. To compare this implicit reward with that of DPO, we calculated it for both methods on the UltraFeedback test set. The results in Figure \ref{fig:reward_modelleing_tpo_dpo} shows that TPO outperforms DPO across various data sizes. These findings suggest that the implicit reward in TPO has a stronger effect compared to DPO. Additionally, it is evident that the average of \( \log \pi_\theta(y|x) \) significantly influences the win rate, particularly with larger dataset sizes.

\subsection{Robustness of TPO and DPO}
\label{sec:robustness_dpo_tpo}
In this section, we investigate the robustness of TPO and DPO across different scenarios. First, we analyze the effect of judgment noise in preference datasets. Next, we explore conditions closer to real-world situations, where two out of three responses generated for the same prompt receive identical scores from a Judger model. Finally, we assess the performance of DPO when initialized from a more robust SFT checkpoint.

\paragraph{TPO displays stronger robustness to judgment noise than DPO.} To compare TPO and DPO under noisy conditions, we use the 40,000 Base setting. We assume that judgment noise exists in the preference dataset, while the gold dataset remains clean. The SFT checkpoint for DPO is fine-tuned on this clean gold dataset. Noise is introduced into 50\% and 100\% of the training set of the preference dataset. For the 50\% noise condition, we randomly select half of the training data and swap the preferred and rejected responses. In the 100\% noise condition, all preferred and rejected responses are completely swapped.

\begin{figure}[!h]
    \centering
    \includegraphics[width=1\linewidth]{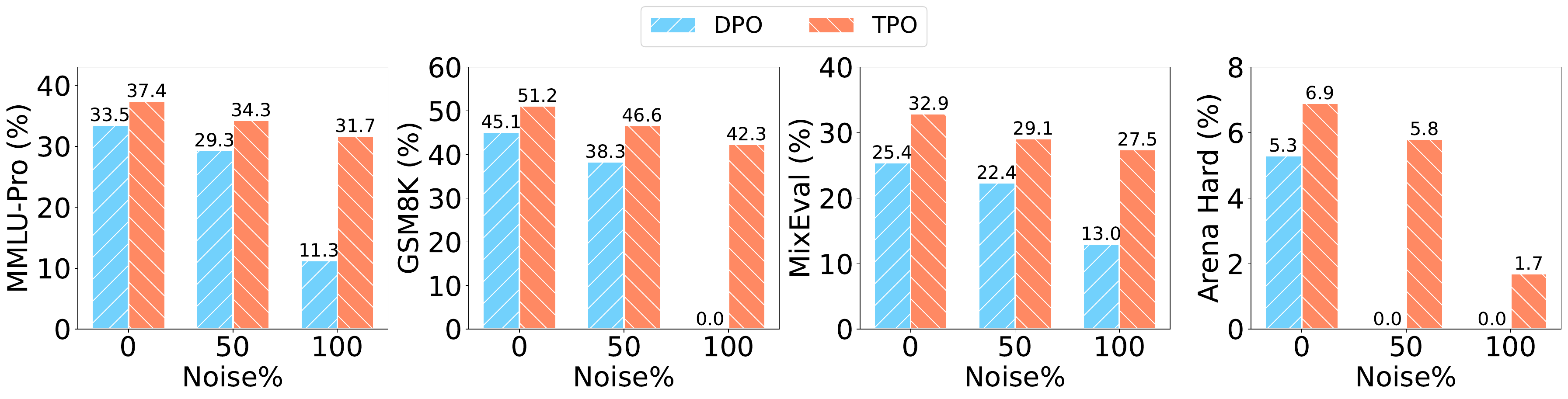}
    \caption{Robustness comparison of TPO and DPO under different percentages of judgment noise in the training set on 40,000 Base setting.}
    \label{fig:noise_analysis}
\end{figure}

The results in Figure \ref{fig:noise_analysis} show that as noise increases, the DPO checkpoint collapses, with model performance on GSM8K and Arena-Hard dropping to zero under 100\% noise. In contrast, TPO maintains acceptable performance even under 100\% noise, although there is a decline across benchmarks. This robustness can be attributed to the \( \log \pi_\theta(y_\text{gold}|x) \) term in TPO, which relies on the gold response. When the preference dataset is noisy, this term helps the model retain acceptable performance. These findings highlight the importance of the NLL loss on the gold response during preference optimization, showcasing its role in mitigating the impact of noise in the preference dataset.

\input{tables/preference_analysis_tpo}

\paragraph{TPO performs reliably across different hypotheses.} TPO is based on the assumption that the dataset \( D = \{x, y_{\text{gold}}, y_w, y_l\} \) satisfies \( y_{\text{gold}} \succ y_w \succ y_l \). However, constructing such a dataset can be challenging, as in many cases, two of the preferences may have very similar scores. To address this, we designed an experiment to evaluate the performance of TPO in a scenario where the Judger model assigns the same score to \( y_{\text{gold}} \) and \( y_w \).  For this purpose, we prepared a dataset of 10,000 examples based on the overall scores in UltraFeedback, ensuring \( S(y_{\text{gold}}) = S(y_w) \). In this setting, the gold and preferred responses share the same overall score, with a 0.5-point difference between them and the rejected response. The results in Table \ref{tab:preference-analysis-tpo} indicate that TPO performs comparably in this new dataset to its performance under the original assumption. This finding suggests that even when \( y_{\text{gold}} \) and \( y_w \) share the same score, using distinct responses for \( y_{\text{gold}} \) and \( y_w \) yields significantly better performance compared to the case where \( y_{\text{gold}} = y_w \), which is similar to the CPO loss function.

\input{tables/sft_analysis_dpo}

\paragraph{TPO is more effective than DPO while using less data.} The recent study \citep{tunstall2023zephyr, saeidi2024insights} shows that the SFT checkpoint is a crucial component for direct preference optimization. Since SFT is a separate process, it can be fine-tuned with additional data to potentially improve performance. In this study, we evaluate the performance of DPO using different SFT checkpoints fine-tuned on larger datasets. Specifically, we use the UltraFeedback 10,000 Base setting and modify the DPO model by replacing the SFT checkpoint with versions fine-tuned on 20,000 and 40,000 data points. The results in Table \ref{tab:sft-analysis-dpo} indicate that DPO fine-tuned on larger SFT datasets achieves improved performance, particularly on Arena-Hard. However, when the SFT checkpoint was fine-tuned on 40,000 data points, DPO's performance on reasoning benchmarks declined compared to the 40,000 Base setting. Additionally, despite utilizing twice as much data as TPO and TPO-L, DPO struggled to outperform them, except on Arena-Hard. This finding underscores that TPO achieves comparable performance to DPO with significantly less data, highlighting its efficiency.


%% file: tables/main_instruct_results.tex
\begin{wraptable}{r}{.55\linewidth}
\vspace{-1.4em}
\small
\centering
\caption{Results for Are-Hard, MT-Bench, MixEval-Hard, MMLU-Pro, MMLU, and GSM8K under the Instruct setting, using off-the-shelf models as the SFT model. Front of the TPO and TPO-L scores, we identify the improvement compared with DPO.}
\label{tab:main_instruct_res}
\resizebox{\linewidth}{!}{%
\begin{tabular}{lcccccc}
\toprule
\multirow{3}{*}{\textbf{Method}} & \multicolumn{6}{c}{\textbf{LLaMA-3B-Instruct}} \\ 
\cmidrule(lr){2-7}
& \multicolumn{3}{c}{\textbf{Reasoning}} & \multicolumn{3}{c}{\textbf{Instruction Following}} \\
\cmidrule(lr){2-4} \cmidrule(lr){5-7}
& {\scriptsize \bf GSM8K} & {\scriptsize \bf MMLU-Pro} & {\scriptsize \bf MMLU} & {\scriptsize \bf MT-Bench} & {\scriptsize \bf Arena-Hard} & {\scriptsize \bf MixEval-Hard} \\
\midrule
SFT\footnotemark[1] & 68.7 & - & 67.1 & 8.1 & - & - \\
\midrule
DPO\footnotemark[1] & 54.5 & 43.7 & \textbf{67.3} & 7.7 & 40.9 & 41.6 \\
CPO\footnotemark[1] & 67.8 & 42.6 & 66.9 & 7.9 & 35.6 & 41.3 \\
IPO\footnotemark[1] & 56.3 & 42.8 & \underline{67.3} & 7.9 & 40.1 & 44.0 \\
ORPO\footnotemark[1] & 63.4 & 42.7 & 66.8 & 7.9 & 34.8 & 36.4 \\
KTO\footnotemark[1] & 66.4 & 42.1 & 67.2 & 8.1 & 34.7 & \underline{44.1} \\
SIMPO\footnotemark[1] & 55.6 & 40.5 & 66.5 & 7.7 & 35.1 & \textbf{45.0} \\
SLIC-HF\footnotemark[1] & 68.2 & 40.2 & 66.9 & 7.7 & 28.6 & 40.0 \\
\midrule
$\text{TPO}_{y2}$ & 77.2 {\raisebox{-0.1em}{\scalebox{0.7}{\textcolor[HTML]{4CC417}{\textbf{(+22.7)}}}}} & \textbf{44.4} {\raisebox{-0.1em}{\scalebox{0.7}{\textcolor[HTML]{4CC417}{\textbf{(+0.7)}}}}} & 65.9 {\raisebox{-0.1em}{\scalebox{0.7}{\textcolor[HTML]{F62217}{\textbf{(-1.4)}}}}} & \textbf{8.2} {\raisebox{-0.1em}{\scalebox{0.7}{\textcolor[HTML]{4CC417}{\textbf{(+0.5)}}}}} & \underline{42.0} {\raisebox{-0.1em}{\scalebox{0.7}{\textcolor[HTML]{4CC417}{\textbf{(+1.1)}}}}} & 40.4 {\raisebox{-0.1em}{\scalebox{0.7}{\textcolor[HTML]{F62217}{\textbf{(-1.2)}}}}} \\
$\text{TPO}_{y3}$ & \underline{77.8} {\raisebox{-0.1em}{\scalebox{0.7}{\textcolor[HTML]{4CC417}{\textbf{(+23.3)}}}}} & \underline{43.8} {\raisebox{-0.1em}{\scalebox{0.7}{\textcolor[HTML]{4CC417}{\textbf{(+0.1)}}}}} & 65.5 {\raisebox{-0.1em}{\scalebox{0.7}{\textcolor[HTML]{F62217}{\textbf{(-1.8)}}}}} & \textbf{8.2} {\raisebox{-0.1em}{\scalebox{0.7}{\textcolor[HTML]{4CC417}{\textbf{(+0.5)}}}}} & \textbf{42.4} {\raisebox{-0.1em}{\scalebox{0.7}{\textcolor[HTML]{4CC417}{\textbf{(+1.5)}}}}} & 41.9 {\raisebox{-0.1em}{\scalebox{0.7}{\textcolor[HTML]{4CC417}{\textbf{(+0.3)}}}}} \\
$\text{TPO}_{y4}$ & \textbf{78.2} {\raisebox{-0.1em}{\scalebox{0.7}{\textcolor[HTML]{4CC417}{\textbf{(+23.7)}}}}} & 43.4 {\raisebox{-0.1em}{\scalebox{0.7}{\textcolor[HTML]{F62217}{\textbf{(-0.3)}}}}} & 65.7 {\raisebox{-0.1em}{\scalebox{0.7}{\textcolor[HTML]{F62217}{\textbf{(-1.6)}}}}} & \underline{8.1} {\raisebox{-0.1em}{\scalebox{0.7}{\textcolor[HTML]{4CC417}{\textbf{(+0.4)}}}}} & 38.9 {\raisebox{-0.1em}{\scalebox{0.7}{\textcolor[HTML]{F62217}{\textbf{(-2.0)}}}}} & 38.6 {\raisebox{-0.1em}{\scalebox{0.7}{\textcolor[HTML]{F62217}{\textbf{(-3.0)}}}}} \\
$\text{TPO-L}_{y2}$ & 77.3 {\raisebox{-0.1em}{\scalebox{0.7}{\textcolor[HTML]{4CC417}{\textbf{(+22.8)}}}}} & 43.7 {\raisebox{-0.1em}{\scalebox{0.7}{\textcolor[HTML]{4CC417}{\textbf{(0.0)}}}}} & 65.7 {\raisebox{-0.1em}{\scalebox{0.7}{\textcolor[HTML]{F62217}{\textbf{(-1.6)}}}}} & \textbf{8.2} {\raisebox{-0.1em}{\scalebox{0.7}{\textcolor[HTML]{4CC417}{\textbf{(+0.5)}}}}} & 39.4 {\raisebox{-0.1em}{\scalebox{0.7}{\textcolor[HTML]{F62217}{\textbf{(-1.5)}}}}} & 39.4 {\raisebox{-0.1em}{\scalebox{0.7}{\textcolor[HTML]{F62217}{\textbf{(-1.2)}}}}} \\
\bottomrule
\end{tabular}
}
\end{wraptable}

%% file: tables/preference_analysis_tpo.tex
\begin{table}[h]
    \caption{Comparison of TPO under a different assumption where the gold and preferred responses differ but have the same quality score.}
    \centering
    \resizebox{\linewidth}{!}{
    \begin{tabular}{l|c|cccccc}
        \toprule
        \textbf{Method} & \textbf{Quality Comparison} & \textbf{GSM8k} & \textbf{MMLU-Pro} & \textbf{MMLU} & \textbf{MT-Bench} & \textbf{Arena-Hard (Avg. \# Token Length)} & \textbf{MixEval-Hard} \\
        \midrule
        
        TPO & $y_{gold} \succ y_w \succ y_l$  & \textbf{52.2} & 37.8 & \textbf{65.3} & \textbf{6.7} & 6.7 (683) & 31.2 \\
        TPO & $y_{gold} \simeq y_w \succ y_l$ & 51.5 & \textbf{37.8} & 65.3 & 6.4 & \textbf{8.2 (642)} & \textbf{32.0} \\

        \bottomrule
    \end{tabular}
    }
    \label{tab:preference-analysis-tpo}
\end{table}

%% file: tables/sft_analysis_dpo.tex
\begin{table}[h]
    \caption{Comparison of TPO and TPO-L with DPO, which uses more robust SFT checkpoints.}
    \centering
    \resizebox{\linewidth}{!}{
    \begin{tabular}{l|cc|cccccc}
        \toprule
        \textbf{Method} & \textbf{SFT Data} & \textbf{Preference Data} & \textbf{GSM8k} & \textbf{MMLU-Pro} & \textbf{MMLU} & \textbf{MT-Bench} & \textbf{Arena-Hard} & \textbf{MixEval-Hard} \\
        \midrule
        
        DPO & 10K ($y_{gold}$) & 10K ($y_w, y_l$) & 36.7 & 28.8 & 59.4 & 6.3 & 6.1 & 25.9\\
        DPO & 20K ($y_{gold}$) & 10K ($y_w, y_l$) & 49.8 & 35.8 & 63.0 & 6.6 & 7.1 & 28.5 \\
        DPO & 40K ($y_{gold}$) & 10K ($y_w, y_l$) & 46.8 & 32.3 & 62.3 & 6.6 & \textbf{7.8} & 25.2 \\
        \midrule
        TPO & none & 10K ($y_{gold}, y_w, y_l$) & \textbf{52.2} & \underline{37.8} & \underline{65.3} & \underline{6.7} & 6.7 & \underline{31.2} \\
        TPO-L & none & 10K ($y_{gold}, y_w, y_l$) & \underline{51.9} & \textbf{38.0} & \textbf{65.4} & \textbf{6.9} & \underline{7.3} & \textbf{33.7} \\

        \bottomrule
    \end{tabular}
    }
    \label{tab:sft-analysis-dpo}
\end{table}

%% file: text/2_related_work.tex
\section{Related Works}
\paragraph{Reinforcement learning from human feedback.} 
RLHF is used to align large language models with human preferences and values~\citep{christiano2017deep, ziegler2019fine, Ouyang2022TrainingLM, bai2022training}. A classical RLHF setting, involves three stages: supervised fine-tuning stage~\citep{zhou2024lima,taori2023stanford, geng2023koala, DatabricksBlog2023DollyV2, kopf2024openassistant, Ding2023EnhancingCL, wang2024openchat, chen2024alpagasus, xia2024less}, reward modeling stage~\citep{gao2023scaling,luo2023wizardmath, chen2024odin,lightman2023let,havrilla2024glore, lambert2024rewardbench}, and policy optimization stage~\citep{schulman2017proximal, anthony2017thinking}. During the final stage, Proximal Policy Optimization (PPO) \citep{schulman2017proximal} is more commonly used. This approach involves optimizing for maximum reward by interacting with a reward model trained using the Bradley-Terry objective. The application of RLHF framework trancends various applications, such as ensuring safety~\citep{dai2023safe}, enhancing helpfulness~\citep{tian2024finetuning,Wang2024ArithmeticCO}, mitigating toxicity~\citep{amini2024direct,korbak2023pretraining,Zheng2023ClickCT}, searching and navigating the web~\citep{nakano2021webgpt}, and improving model reasoning abilities~\citep{havrilla2024teaching}. 
While Reinforcement Learning with Human Feedback (RLHF) enhances model performance, it encounters challenges such as the collection of preference data \citep{casper2023open}, unstable training processes, and the generation of biased or overly verbose responses \citep{dubois2024length,singhal2023long,wang2023far}.

\paragraph{Offline vs. iterative preference optimization.} 
Owing to the challenges and complexity of online preference optimization techniques\citep{zheng2023secrets, santacroce2023efficient}, recent works have proposed simple and efficient offline algorithms. One among them is Direct Preference Optimization (DPO)\citep{Rafailov2023DirectPO}. It makes use of an implicit reward which is in terms of the parameterized policy and a reference policy. Due to the absence of an explicit reward model, DPO is limited in its ability to sample preferences from the optimal policy. Addressing this, works like \citep{Zhao2023SLiCHFSL, liu2024statistical} explore augmenting preference data using a trained or refined SFT policy. Iterative training methods which continuously update the reference model with the most recent policy model or generate new preference pairs at each iteration have been explored by works \citep{dong2024rlhf, Kim2024sDPODU, Rosset2024DirectNO, xiong2024iterative,yuan2024self}. In this work, we experiment in an offline setting, without any iterative training.

\paragraph{Preference optimization objectives.} Besides DPO, various preference optimization algorithms have been proposed.A variety of preference optimization objectives have been proposed besides DPO~\citep{dong2023raft, liu2024lipo, song2024preference, yuan2023rrhf}. However, this family of methods either relies on explicit rewards during preference optimization or demands a fine-tuned reference model. In contrast, TPO avoids this by using three preferences. Similar to TPO, simpler objectives that do not rely on a reference model have been proposed\citep{Hong2024ORPOMP, xu2023some}. In this work, we compare TPO to a series of offline algorithms, including RRHF \cite{yuan2023rrhf}, SLiC-HF \cite{Zhao2023SLiCHFSL}, DPO \cite{Rafailov2023DirectPO}, IPO \cite{Azar2023AGT}, CPO \cite{xu2024contrastive}, KTO \cite{Ethayarajh2024KTOMA}, ORPO \cite{Hong2024ORPOMP}, and SimPO \cite{meng2024simpo}, and find that TPO can outperform them in both efficiency and performance. Similar to TPO, SimPO does not impose KL regularization, however, the robustness of its effective learning is dependent on various factors like preference datasets with diverse domain coverage. Whereas, TPO is not constrained by this.

%% file: text/6_conclusion.tex
\section{Conclusion}
In this study, we introduced the Triple Preference Optimization (TPO) method, a one-step preference optimization approach that enhances both the instruction-following and reasoning abilities of the policy model. We developed TPO-L, a length-controlled variant, by defining a reward margin and replacing summation with the average response sequence. Our results show that TPO and TPO-L outperform existing methods across benchmarks, effectively address the verbosity problem, and are less sensitive to dataset size, particularly in the case of TPO-L. Additionally, TPO demonstrates greater robustness to noisy data and performs better on less data than DPO.  A detailed discussion of the limitations can be found in Appendix \ref{sec:app_limitation}.

%% file: text/7_appendix.tex
\clearpage
\appendix

\renewcommand{\thelemma}{\arabic{lemma}} 
\renewcommand{\thetheorem}{\arabic{theorem}} 

\section*{Appendix}
\section{Limitations and Future works}
\label{sec:app_limitation}
\paragraph{Exploring the Quality Margin Between Responses.} The main hypothesis of TPO focuses on maintaining a margin difference between gold and preferred responses. Although we evaluated TPO in scenarios where the gold and preferred responses share the same quality score, further exploration of the margin between gold and preferred responses, as well as between preferred and rejected responses, is still needed. Our analysis of the UltraFeedback dataset indicates that increasing the margin between gold and preferred responses significantly reduces the number of qualifying samples, leaving only 10,000 that meet the conditions. Therefore, a more in-depth analysis of this area is necessary.

\paragraph{Exploration on Reward Margin.} Although TPO-L demonstrated impressive performance across various settings, it requires considerable effort to determine the optimal value for the reward margin \( \gamma \). This limitation restricts the applicability of TPO-L and similar methods, such as SimPO, across different tasks. Developing dynamic approaches to adjust the reward margin automatically based on the specific task could be an interesting direction for future work.

\paragraph{Safety and Honesty.} As discussed in Section \ref{sec:tpo_explanaition}, the TPO loss function incorporates two objective functions. Recent studies \citep{dai2023safe, zhou2023beyond} have explored multi-objective preference optimization methods, which require preparing multiple preference pairs. Safety is an important domain where multiple objectives, such as safety and helpfulness, are equally critical \citep{handa2024jailbreaking}. Exploring the impact of TPO on such multi-objective tasks could provide valuable insights. A key advantage of TPO compared to other methods is that data collection for TPO is significantly easier than for multi-objective methods like DPO.

\paragraph{In-Depth Exploration of Reasoning.} As shown in Tables \ref{tab:main_res} and \ref{tab:downstream_tpo}, TPO and TPO-L achieved strong results on reasoning benchmarks. However, to further demonstrate TPO’s effectiveness on reasoning tasks, we encourage researchers to focus on reasoning-specific datasets such as UnSeenTimeQA \citep{uddin2024unseentimeqa} and explore preference optimization designed for reasoning tasks. Developing and evaluating reasoning-based preference datasets would provide deeper insights into TPO’s capabilities.

\paragraph{Online Learning.} Recent study \citep{park2024disentangling} suggests that online versions of DPO address the overfitting challenges associated with offline versions, often achieving better performance. Exploring this area could be particularly valuable, as superior results on datasets like UltraFeedback-ArmoRM—which closely align with the concept of online policy learning—support the feasibility of online learning approaches.

\setcounter{equation}{0}
\section{Derivation}

\subsection{Deriving the optimal policy under the Preference Objective}
\label{sec:appendix_optimal}
In this section, we derive the optimal policy achieved by optimizing the objective in Equation \ref{rl_obj_r} from Section \ref{sec:deriving_TPO}. For a given prompt $x$, the objective can be analogously written as follows:

\begin{equation*}
\begin{split}
\max_{\pi} \;\mathbb{E}_{y \sim \pi(y|x)} \left[r(x,y)-\beta\log\pi(y|x)  \right]
s.t. \sum_{y}\pi(y|x) = 1
\end{split}
\label{eq:1a}
\end{equation*}

Next, we form a lagrangian for the above objective with $\lambda$ being the lagrangian multiplier.

\begin{equation*}
\begin{split}
\mathcal{L} = \sum_{y}\pi(y|x)r(x,y) - \beta\bigg[\sum_{y}\pi(y|x)\log\pi(y|x)\bigg] -\lambda\bigg[1-\sum_{y}\pi(y|x)\bigg]
\end{split}
\label{eq:2a}
\end{equation*}

Differentiating $\mathcal{L}$ with respect to $\pi(y|x)$ results in,

\begin{equation*}
\begin{split}
\frac{\partial\mathcal{L}}{\partial_{\pi(y|x)}} = r(x,y) - \beta\bigg[\log\pi(y|x) + 1\bigg] - \lambda
\end{split}
\label{eq:3a}
\end{equation*}

To obtain the optimal policy, we can set the above equation to zero and solve for $\pi(y|x)$.

\begin{equation*}
\begin{split}
r(x,y) - \beta\bigg[\log\pi(y|x) + 1\bigg] - \lambda = 0
\end{split}
\label{eq:4a}
\end{equation*}

\begin{equation*}
\begin{split}
\log\pi(y|x) = \frac{1}{\beta}r(x, y) -\frac{\lambda}{\beta} - 1
\end{split}
\label{eq:5a}
\end{equation*}

\begin{equation*}
\begin{split}
\pi(y|x) = \exp{(\frac{1}{\beta}r(x, y))}.\exp{(\frac{-\lambda}{\beta}-1)}
\end{split}
\label{eq:6a}
\end{equation*}

Since $\sum_{y}\pi(y|x) = 1$, the second exponent is a partition function that does normalization as shown below:

\begin{equation*}
\begin{split}
\bigg[\sum_{y} \exp{(\frac{1}{\beta}r(x, y))}\bigg].\exp{(\frac{-\lambda}{\beta}-1)} = 1
\end{split}
\label{eq:7a}
\end{equation*}

\begin{equation*}
\begin{split}
\exp{(\frac{-\lambda}{\beta}-1)} = \bigg[\sum_{y} \exp{(\frac{1}{\beta}r(x, y))}\bigg]^{-1}
\end{split}
\label{eq:8a}
\end{equation*}

Hence, the partition function $Z(x) = \sum_{y} \exp{(\frac{1}{\beta}r(x, y))}$ and the optimal policy $\pi_{r}(y|x)$ induced by reward function $r(x, y)$  is therefore given by,

\begin{equation}
\begin{split}
\pi_{r}(y|x) = \frac{1}{Z(x)}\exp{(\frac{1}{\beta}r(x, y))}
\end{split}
\label{eq:9a}
\end{equation}

Now, we can express the reward function in terms of an optimal policy $\pi_{r}$ by performing some algebraic transformations on Equation \ref{eq:9a} as shown below,

\begin{equation*}
\begin{split}
\pi_{r}(y|x).Z(x) = \exp{(\frac{1}{\beta}r(x, y))}
\end{split}
\label{eq:10a}
\end{equation*}

Taking logarithm and multiplying by $\beta$ on both sides,

\begin{equation}
\begin{split}
r(x, y) = \beta\log\pi_{r}(y|x) + \beta\log Z(x)
\end{split}
\label{eq:11a}
\end{equation}

\subsection{Deriving the Gradient of the TPO Objective}
\label{sec:gradient_tpo}
\setcounter{equation}{0}

In this section, we derive the gradient of the TPO objective:
\begin{equation}
\nabla_{\theta} \mathcal{L}_{\text{TPO}} = -\nabla_{\theta}\mathds{E}_{(x,y_{ref},y_{w},y_{l})\sim\mathcal{D}}\;[\; \alpha \log\pi_{\theta}(y_{ref}|x) + \log\sigma(\beta\log\pi_{\theta}(y_{w}|x) - \beta\log\pi_{\theta}(y_{l}|x)) \;]
\label{eq:1b}
\end{equation}
We can rewrite the RHS of the Equation~\ref{eq:1b} as
\begin{equation}
\nabla_{\theta} \mathcal{L}_{\text{TPO}} = -\mathds{E}_{(x,y_{ref},y_{w},y_{l})\sim\mathcal{D}}\;[\; \underbrace{\alpha\nabla_{\theta} \log\pi_{\theta}(y_{ref}|x)}_{\text{(a)}} + \underbrace{\nabla_{\theta}\log\sigma(\beta\log\pi_{\theta}(y_{w}|x) - \beta\log\pi_{\theta}(y_{l}|x))}_{\text{(b)}} \;]
\label{eq:2b}
\end{equation}

In equation~\ref{eq:2b}, the part (b) can be rewritten with
\[ u = \beta\log\pi_{\theta}(y_{w}|x) - \beta\log\pi_{\theta}(y_{l}|x) \]

\begin{equation*}
\nabla_{\theta}\log\sigma(u) = \frac{1}{\sigma(u)}\nabla_{\theta}\sigma(u)
\label{eq:3b}
\end{equation*}

\begin{equation*}
\nabla_{\theta}\log\sigma(u) = \frac{\sigma^{'}(u)}{\sigma(u)}\nabla_{\theta}(u)
\label{eq:4b}
\end{equation*}

Using the properties of sigmoid function function $\sigma^{'}(u) = \sigma(u)(1-\sigma(u)$ and $\sigma(-u) = 1 - \sigma(u)$,

\begin{equation*}
\nabla_{\theta}\log\sigma(u) = \frac{\sigma(u)(1 - \sigma(u))}{\sigma(u)}\nabla_{\theta}(u)
\label{eq:5b}
\end{equation*}

\begin{equation*}
\nabla_{\theta}\log\sigma(u) = (1 - \sigma(u))\nabla_{\theta}(u)
\label{eq:6b}
\end{equation*}

\begin{equation*}
\nabla_{\theta}\log\sigma(u) = \sigma(-u)\nabla_{\theta}(u)
\label{eq:7b}
\end{equation*}

\begin{equation}
    \nabla_{\theta}\log\sigma(u) = \beta\sigma(\beta\log\pi_{\theta}(y_{l}|x) -\beta\log\pi_{\theta}(y_{w}|x))\;[\nabla_{\theta}\log\pi(y_{w}|x)-\nabla_{\theta}\log\pi(y_{l}|x)]
\label{eq:8b}
\end{equation}

Plugging Equation~\ref{eq:8b} into Equation~\ref{eq:2b} we get,
\begin{align}
\nabla_{\theta} \mathcal{L}_{\text{TPO}} = & -\mathds{E}_{(x,y_{ref},y_{w},y_{l})\sim\mathcal{D}}\;[\alpha\nabla_{\theta}\log\pi(y_{ref}|x) \nonumber \\
& + \beta\sigma(\beta\log\pi_{\theta}(y_{l}|x) - \beta\log\pi_{\theta}(y_{w}|x)) \nonumber \\
& \times [\nabla_{\theta}\log\pi(y_{w}|x)-\nabla_{\theta}\log\pi(y_{l}|x)] ]
\end{align}

\subsection{Theory Behind TPO}
\label{sec:theory_TPO}
In this section, we provide a theoretical foundation for the TPO algorithm, drawing inspiration from~\citet{rafailov2024direct}. We observe that the preference optimization objective aligns with the principles of a Bradley-Terry model, where the reward parameterization is defined as $r^{}(x, y) = \beta\log\pi^{}_{\theta}(y|x)$. Consequently, we optimize our parametric model $\pi_{\theta}$ in a manner similar to reward model optimization, as shown by \citet{ouyang2022training}. 
We expand on the theory underlying this reparameterization of the reward function, illustrating that it does not constrain the range of reward models that can be modeled and ensures accurate retrieval of the optimal policy. We initiate this discussion by following the insights presented in DPO about the equivalent class of reward models.

\paragraph{Definition 1.} \textit{Two reward functions $r(x,y)$ and $r^{'}(x,y)$ are equivalent iff $r(x,y)-r^{'}(x,y) = g(x)$ for some function $g$.}
\\
\\
We can state the following two lemmas as it is apparent that there exists an equivalence relation, dividing the set of reward functions into distinct classes.

\begin{lemma}
\textit{Under the Plackett-Luce, and in particular the Bradley-Terry preference framework, two reward functions from the same class induce the same preference distribution.} \citep{rafailov2024direct}
\end{lemma}

\begin{lemma}
\textit{Two reward functions from the same equivalence class induce the same optimal policy under the constrained RL problem.} \citep{rafailov2024direct}
\end{lemma}
The proofs are shown in Appendix \ref{lemmas}.

\begin{theorem}\label{thm:theorem3.1}
\textit{Under mild assumptions, all reward classes consistent with Plackett-Luce models can be represented with the reparameterization \(r(x, y) = \beta\log\pi(y|x)\) for some model \(\pi(y|x)\).} \citep{rafailov2024direct}
\end{theorem}

As proposed in DPO, upon imposing certain constraints on the under-constrained Plackett-Luce family of preference models, such that we preserve the class of representable reward model, it possible to explicitly make the optimal policy in Equation \ref{optimal_policy} from Section \ref{sec:deriving_TPO} analytically tractable for all prompts $x$. The theorem is elaborated in Appendix \ref{theorem}. We further elaborate our theoretical basis for defining and optimally addressing the TPO objective within a multi-objective optimization framework.

\phantomsection
\label{def:pareto_optimal}
\paragraph{Definition 2.} \textit{Let $f_{i}$ denote $i^{th}$ objective, $\mathcal{S}$ denote the feasible policy space, then in a multi-objective optimization setting, a policy $\pi^{\ast}\in \mathcal{S}$ is said to be Pareto optimal if there does not exist another policy $\pi\in \mathcal{S}$ such that $f_{i}(\pi) \leq f_{i}(\pi^{\ast})$ for all $i = 1,...,k$ and $f_{j}(\pi) < f_{j}(\pi^{\ast})$ for at least one index j.}
\\
\\
Looking at the objectives in Equation \ref{tpo_preference} and Equation \ref{tpo_sft} from Section \ref{sec:deriving_TPO}, it is obvious that optimizing them together is non-trivial; that is, there does exist a policy that is optimal with respect to both objectives. It can be seen that the objectives are conflicting with each other, especially when $y_{\text{gold}}\sim y_{w}$, as one objective is maximizing the log probability and the other is minimizing the log probability. This means that the objectives are at least partly conflicting. For a multi-objective problem, \citet{miettinen1999nonlinear} show that optimizing one objective and converting the other objective/s as a constraint with an upper bound, the solution to this $\epsilon-constrained$ problem is Pareto optimal. This shows that optimizing the TPO objective, which is a bi-objective problem, gives an optimal policy that is Pareto optimal as defined in \ref{def:pareto_optimal}.

\subsection{Proof of Lemma}
\label{lemmas}
In this section, we will prove the lemmas from Section \ref{sec:theory_TPO}.

\paragraph{Lemma 1 Restated.} Under the Plackett-Luce preference framework, and in particular the Bradley-Terry framework, two reward functions from the same equivalence class induce the same preference distribution.

$Proof.$ Let's consider two reward functions, $r(x,y)$ and $r'(x,y)$. They are said to be equivalent if they can be related by $r'(x,y) = r(x,y) + g(x)$ for some function $g$. We analyze this in the context of the general Plackett-Luce model, which includes the Bradley-Terry model (special case when $K=2$). Here, we denote the probability distribution over rankings generated by a given reward function $r(x,y)$ as $p_r$. Given any prompt $x$, responses $y_1, ..., y_K$, and a ranking $\tau$, we can establish the following:

\begin{align*}
p_{r'}(\tau \mid y_1, \ldots, y_K, x) &= \prod_{k=1}^K \frac{\exp(r'(x, y_{\tau(k)}))}{\sum_{j=k}^K \exp(r'(x, y_{\tau(j)}))}\\
&= \prod_{k=1}^K \frac{\exp(r(x, y_{\tau(k)}) + g(x))}{\sum_{j=k}^K \exp(r(x, y_{\tau(j)}) + g(x))}\\
&= \prod_{k=1}^K \frac{\exp(g(x)) \exp(r(x, y_{\tau(k)}))}{\exp(g(x)) \sum_{j=k}^K \exp(r(x, y_{\tau(j)}))}\\
&= \prod_{k=1}^K \frac{\exp(r(x, y_{\tau(k)}))}{\sum_{j=k}^K \exp(r(x, y_{\tau(j)}))}\\
&= p_r(\tau \mid y_1, \ldots, y_K, x),
\end{align*}
This completes the proof.

\paragraph{Lemma 2 Restated.}Two reward functions from the same equivalence class induce the same optimal policy under the constrained RL problem.

$Proof.$ Let's consider two reward functions, $r(x,y)$ and $r'(x,y)$. They are said to be equivalent if they can be related by $r'(x,y) = r(x,y) + g(x)$ for some function $g$. Let $\pi_{r}$ and $\pi_{r^{'}}$ be the optimal policies induced by their corresponding reward functions. By Equation \ref{optimal_policy} from Section \ref{sec:deriving_TPO}, for all $x,y$ we have,

\[
\begin{aligned}
\pi_{r'}(y \mid x) &= \frac{1}{\sum_y  \exp \left(\frac{1}{\beta} r'(x,y)\right)} \exp \left(\frac{1}{\beta} r'(x,y)\right) \\
&= \frac{1}{\sum_y  \exp \left(\frac{1}{\beta} (r(x,y) + g(x))\right)} \exp \left(\frac{1}{\beta} \big(r(x,y) + g(x)\big)\right) \\
&= \frac{1}{\exp \left(\frac{1}{\beta} g(x)\right)\sum_y \exp \left(\frac{1}{\beta} r(x,y)\right)} \exp \left(\frac{1}{\beta} r(x,y)\right)\exp \left(\frac{1}{\beta} g(x)\right) \\
&= \frac{1}{\sum_y \exp \left(\frac{1}{\beta} r(x,y)\right)} \exp \left(\frac{1}{\beta} r(x,y)\right) \\
&= \pi_r(y \mid x),
\end{aligned}
\]
This completes the proof.

\subsection{Proof of Theorem}
\label{theorem}
\setcounter{equation}{0}
\paragraph{Theorem 1 Restated.} \textit{For a parameter $\beta > 0$, all reward equivalence classes can be reparameterized as $r(x, y) =  \beta\log\pi(y|x)$ for some model $\pi(y|x)$.}
\\

$Proof.$ Consider a reward function $r(x, y)$, which induces an optimal model $\pi_{r}(y|x)$ under the MERL framework, which takes the form as shown in Eq.\ref{optimal_policy} from Section \ref{sec:deriving_TPO}. Following, Equation \ref{eq:11a} in Section \ref{sec:appendix_optimal} of Appendix, we have:

\begin{equation}
\begin{split}
r(x, y) = \beta\log\pi_{r}(y|x) + \beta\log Z(x)
\end{split}
\label{eq:1z}
\end{equation}

where $Z(x) = \sum_{y} \exp{(\frac{1}{\beta}r(x, y))}$ is the partition function of the optimal policy induced by the reward function $r(x, y)$. Let $r^{'}(x, y)$ be a new reward function such that $r^{'}(x, y) = r(x, y) - \beta\log Z(x)$. It is obvious that the new reward function is within the equivalence class of $r$, and we have:

\begin{equation*}
\begin{split}
r^{'}(x, y) = r(x, y) - \beta\log Z(x)
\end{split}
\end{equation*}

From the Equation \ref{eq:1z}, we get

\begin{equation*}
\begin{split}
r^{'}(x, y) = \beta\log\pi_{r}(y|x) + \beta\log Z(x) - \beta\log Z(x)
\end{split}
\end{equation*}

\begin{equation*}
\begin{split}
r^{'}(x, y) = \beta\log\pi_{r}(y|x)
\end{split}
\end{equation*}

This completes the proof.

\paragraph{Proposition 1.} For a parameter $\beta > 0$, every equivalence class of reward functions has a unique reward function $r(x, y)$, which can be reparameterized as $r(x, y) = \beta\log\pi(y|x)$ for some model $\pi(y|x)$.

$Proof-by-Contradiction.$ Let us assume that we have two reward functions from the same class, such that $r^{'}(x, y) = r(x, y) + g(x)$. Assume that $r^{'}(x, y) = \beta\log\pi^{'}(y|x)$ for some model $\pi^{'}(y|x)$ and $r(x, y) = \beta\log\pi(y|x)$ for some model $\pi(y|x)$, such that $\pi^{'} \neq \pi$. We then have,
\[
\begin{aligned}
r^{'}(x, y) &= r(x, y) + g(x) \\
&= \beta\log\pi(y|x) + g(x) \\
&= \beta\log\pi(y|x) + \beta\log\exp{(\frac{1}{\beta}g(x))} \\
&= \beta\log\pi(y|x)\exp{(\frac{1}{\beta}g(x))}\\
&= \beta\log\pi^{'}(y|x)
\end{aligned}
\]
for all prompts \textit{x} and completions \textit{y}. Then, we must have $\pi(y|x)\exp{(\frac{1}{\beta}g(x))} = \pi^{'}(y|x)$. Since these are probability distributions, summing over \textit{y} on both sides,

\[
\begin{aligned}
\sum_{y}\big[\pi(y|x)\exp{(\frac{1}{\beta}g(x))}\big] &= \sum_{y}\pi^{'}(y|x) \\
\exp{(\frac{1}{\beta}g(x))} &= 1 \\
\end{aligned}
\]

Since $\beta > 0$, $g(x)$ must be 0 for all $x$. Therefore, we will have $r(x, y) = r^{'}(x, y)$, which contradicts our initial condition of $\pi^{'} \neq \pi$.

Thus, by contradiction, we have shown that every reward class has a unique reward function that can be represented by the reparameterization in Theorem \ref{thm:theorem3.1}.

\section{Implementation Details}
\label{sec:app_implementation_details}
We used Llama-3-8B \footnote{\url{https://huggingface.co/meta-llama/Meta-Llama-3-8B}} for the Base setting, and Llama-3-8B-Instruct \footnote{\url{https://huggingface.co/meta-llama/Meta-Llama-3-8B-Instruct}} for the Instrcut setting. Hyperparameter tuning is essential for optimizing the performance of preference optimization methods. To identify the best hyperparameters, we explored various learning rates [3e-7, 5e-7, 6e-7, 1e-6] and batch sizes [32, 64, 128, 256]. Our observations indicate that preference optimization methods perform best with a batch size of 32 for a training size of 5,000, 32 for 10,000, 64 for 20,000, and 128 for 60,000. However, for large datasets like 60,000, TPO performs best with a batch size of 256. Based on these findings, we fixed these batch sizes for all preference optimization experiments.

\input{tables/tpo_hyperparameters}

Additionally, we set the maximum sequence length to 1024 for Base setting and 2048 for Instruct setting and applied a cosine learning rate schedule with a 10\% warm-up phase for the preference optimization dataset. We followed Table 7 in SimPO \citep{meng2024simpo} for a search on the hyperparameter ranges used for the baseline methods, while Table \ref{tab:tpo_hyper} lists the hyperparameters for TPO and TPO-L under each experimental setting. Moreover, all the training experiments in this paper were conducted on 8×A100 GPUs.

\begin{figure}[h]
    \centering
    \includegraphics[width=1\linewidth]{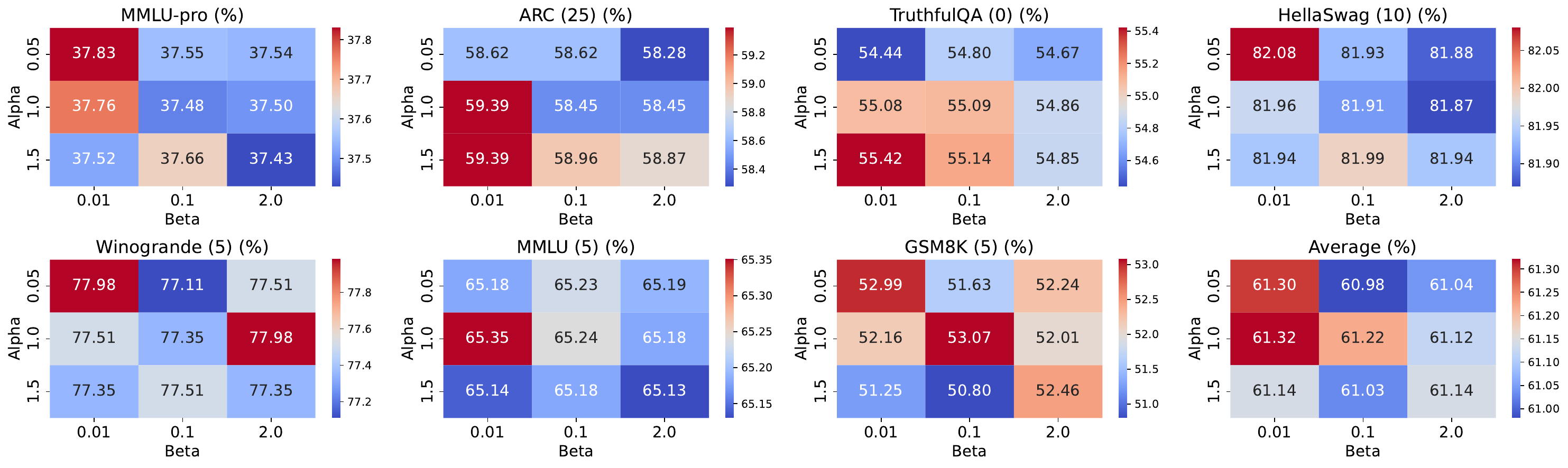}
    \caption{Comparison of TPO on different hyperparameters.}
    \label{fig:ypo_hyper_parameters}
\end{figure}

\paragraph{Exploration on $\alpha$ and $\beta$ in TPO.} As shown in Figure \ref{fig:ypo_hyper_parameters}, we investigated the influence of $\alpha$ and $\beta$ on TPO performance. Our findings indicate that the optimal performance range lies within $\alpha \in [0.05, 1]$ and $\beta \in [0.01, 0.1]$. However, for Instruct models, the best performance was achieved at a lower $\alpha$ value of 0.05. This suggests that the regularization term $\alpha$ plays a crucial role in optimizing different types of models. Given that Instruct models undergo extensive supervised fine-tuning on large instruction datasets, they are already well-regularized. Consequently, a lower regularization strength is necessary to prevent overfitting and to allow the model to learn complex patterns from the data.

\input{tables/impact_ba_acc}

\paragraph{Impact of Batch size per device on Arena-Hard for the instruction-tuned models.} We investigated the performance of TPO under various batch sizes and accumulation configurations while keeping the total batch size fixed at 256 across 8 GPUs. For instance, we experimented with a batch size per device of 2 and gradient accumulation steps of 16 (\textbf{Scenario 1}), as well as a batch size per device of 4 and gradient accumulation sptes of 8 (\textbf{Scenario 2}). Our results, summarized in Table \ref{tab:impact_ba_acc}, indicate that smaller batch sizes per device can negatively impact a model's instruction-following capabilities, specifically on Arena-hard.

\section{Mistral Results}
\label{sec:app_mistral_result}
\footnotetext[1]{We loaded the last checkpoint reported in the SimPO repository for the current methods and evaluated them across different benchmarks.}

\input{tables/mistral_instruction_benchmarks_instruct}

To evaluate the generalizability of the TPO method, we repeated all experiments using Mistral models. For the Base setting, we used Mistral-7B-v0.3 \footnote{\url{https://huggingface.co/mistralai/Mistral-7B-v0.3}} , and for the Instruct setting, we used Mistral-Instruct-v0.2 \footnote{\url{https://huggingface.co/mistralai/Mistral-7B-Instruct-v0.2}}. In the Base setting, we utilized the UltraFeedback dataset, following the same setup described in Section \ref{sec:experiment_setup}. For the Instruct setting, we employed the Mistral-UltraFeedback-PairRM \footnote{\url{https://huggingface.co/datasets/princeton-nlp/mistral-instruct-ultrafeedback}} dataset, which was introduced in the SimPO paper. This dataset was created by generating five different responses per prompt using Mistral-7B-SFT \footnote{\url{https://huggingface.co/mistralai/Mistral-7B-Instruct-v0.2}} and ranking them with PairRM \citep{llm-blender-2023}, a reward model.

\input{tables/mistral_instruction_benchmarks}

The results in Table \ref{tab:mistral_downstream_tpo} indicate that TPO and TPO-L in the Base setting significantly outperform other methods on downstream tasks, particularly GSM8K, MMLU, and MMLU-Pro. Additionally, TPO in the Instruct setting is the only method that achieves better performance on GSM8K compared to the SFT checkpoint. Furthermore, in the Instruct setting, TPO outperforms all other methods in terms of the average accuracy across downstream tasks.

Due to the cost associated with instruction-following benchmarks, we limited our experiments to comparisons with DPO methods in the Base and Instruct settings. The results in Tables \ref{tab:mistral_instruct_res} and \ref{tab:mistral_instruct_instruction_bench} demonstrate that TPO not only surpasses DPO on downstream tasks but also it outperforms DPO on Arena-Hard and MixEval-Hard. These findings highlight the effectiveness of TPO compared to DPO as a primary method across various settings.

\input{tables/mistral_different_chat_template}

We also examined the impact of different chat templates across various benchmarks. The results in Table \ref{tab:mistral_chat_template_analysis} indicate that using the chat completion template designed for Mistral-Base on the Instruction version of Mistral reduces instruction-following performance but positively impacts downstream tasks, particularly GSM8K.

\input{tables/mistral_downstream}

\section{Downstream Task Evaluation}
\label{sec:app_down_stream_tasks}
To further investigate the impact of preference optimization methods, we evaluated their performance on various downstream tasks. Results for MMLU and GSM8K are included in the main text, while additional results for MMLU, ARC, HellaSwag, TruthfulQA, Winograd, and GSM8K are presented in Table \ref{tab:downstream_tpo}. Following established evaluation protocols, we report results for all models. TPO’s primary goal is to enhance a model’s reasoning ability while simultaneously improving its instruction-following performance. As noted in the main text, TPO demonstrated impressive results on both GSM8K and Arena-Hard simultaneously.

The results in Table \ref{tab:downstream_tpo} indicate that TPO and TPO-L also achieved strong performance on ARC and showed notable improvements in the TruthfulQA benchmark. Another significant observation is their performance on HellaSwag, where TPO and TPO-L consistently outperformed other methods across all settings. The performance of TPO and TPO-L differs depending on the dataset size. While their results on smaller datasets, such as 5,000 and 10,000 samples, are slightly smaller than other methods, they outperform the alternatives on larger datasets, including 20,000 and 40,000 samples, as well as in the Instruct settings. These findings further support the conclusion that TPO and TPO-L have a substantial impact on enhancing a model’s performance on reasoning benchmarks.

\input{tables/llama3_downstream}

\section{GPT-4o Judgment for Arena-Hard}
\label{sec:app_gpt_4o}
\footnotetext[1]{We loaded the last checkpoint reported in the SimPO repository for the current methods and evaluated them across different benchmarks. We also used values for downstream tasks reported in the SimPO paper.}
Comparing the Arena-Hard results in the SimPO paper with our study on the same benchmark highlights a significant discrepancy due to differences in the Judgment models used. In Arena-Hard, the default Judger is GPT-4-turbo (gpt-4-1106-preview), but we replaced it with GPT-4o for our evaluations. The primary reason for this change is the improved reliability of GPT-4o, which has shown better performance compared to the earlier version across different benchmarks. Additionally, the lower cost of evaluating each model per judgment was another motivation for selecting GPT-4o.

\begin{figure}[h]
    \centering
    \includegraphics[width=0.65\linewidth]{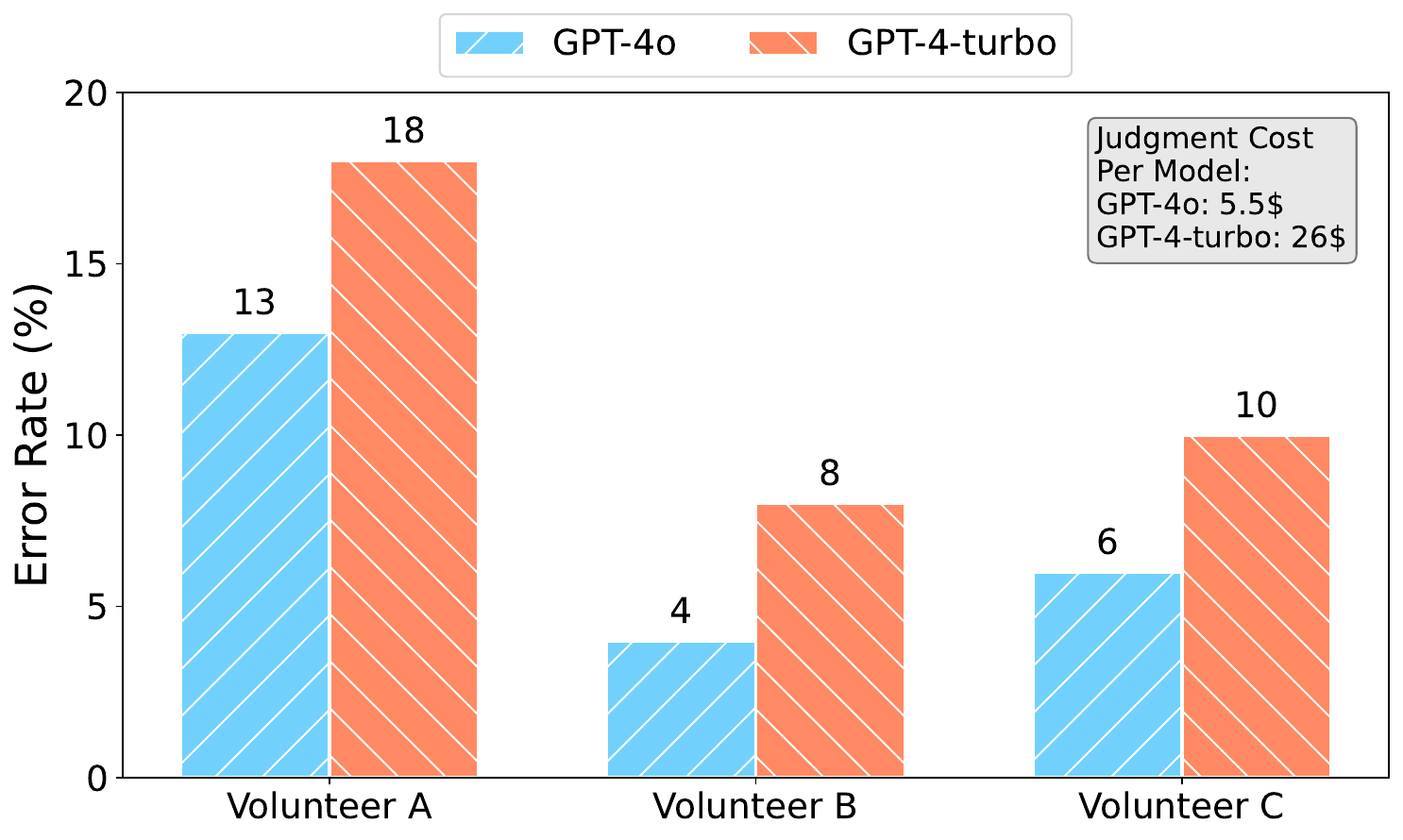}
    \caption{Comparative analysis of GPT-4o and GPT-4-turbo, focusing on error rates in judgment of responses for Arena-Hard and cost per model.}
    \label{fig:comp_gpt_4o_gpt_4_turbo}
\end{figure}

To validate this decision, we conducted an experiment to identify the best model for judgment. Using the same responses generated by the TPO model on the Arena-Hard benchmark, we evaluated them with both GPT-4-turbo and GPT-4o. We then selected 100 samples and hired three volunteer researchers to assess the judgments generated by the two models. The results, presented in Table \ref{fig:comp_gpt_4o_gpt_4_turbo}, show that GPT-4o's judgments are more closely aligned with human decisions compared to GPT-4-turbo. Based on this finding, we evaluated all models in our study using GPT-4o as the Judgment model on Arena-Hard.

\section{Token Length Analysis}
\label{sec:app_token_length_analysis}
In Section \ref{sec:verbosity_problem}, we observed that TPO, even with summation in the loss function, produces shorter responses compared to DPO and, in some settings, compared to SimPO. One possible explanation could be that the gold response used in TPO is shorter in length compared to the preferred and rejected responses. In this section, we analyze the average token length of gold, as well as preferred and rejected responses across different settings.

\begin{figure}[h]
    \centering
    \includegraphics[width=1.0\linewidth]{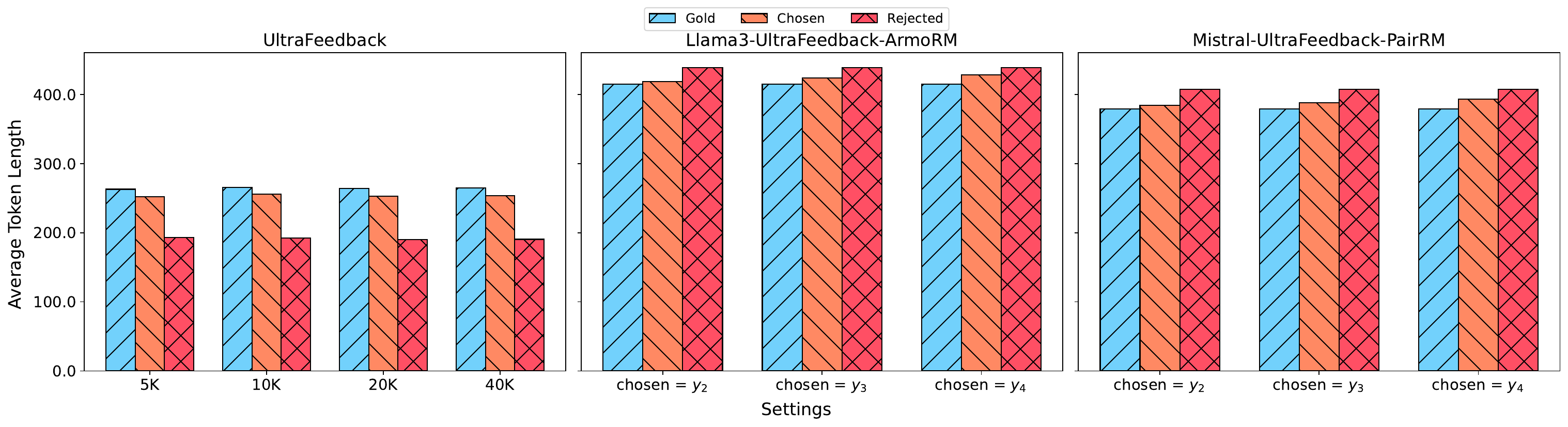}
    \caption{Comparison of token lengths for gold, chosen, and rejected responses across various datasets.}
    \label{fig:datasets_length}
\end{figure}

In the Base setting, models are fine-tuned on UltraFeedback. The token length analysis in Figure \ref{fig:datasets_length} reveals that the average token length for chosen and rejected responses is shorter than that of the gold responses. This observation indicates that current preference optimization methods, even with shorter responses, tend to generate longer outputs compared to TPO checkpoints, as shown in Figure \ref{fig:length_analysis}. 

In the Instruct setting, other models are fine-tuned on gold and rejected responses, while TPO explores performance using different chosen responses. It is worth noting that Llama-UltraFeedback-ArmoRM and Mistral-UltraFeedback-PairRM datasets provide five ranked responses for each prompt. Even in this scenario, TPO generates shorter responses compared to most current preference optimization methods.

In summary, As shown in Figure \ref{fig:datasets_length}, the average token length of gold responses is not shorter than that of chosen or rejected responses. This indicates that the shorter responses generated by TPO, compared to methods like DPO that use summation in their loss function, are due to the influence of the Behavioral Cloning objective on different response types. This differentiation between preferred and rejected responses is an additional advantage of TPO.

%% file: tables/tpo_hyperparameters.tex
\begin{table*}[!h]
\small
\centering
\caption{The hyperparameter values in TPO and TPO-L utilized for each training setting.}
\label{tab:tpo_hyper}
\setlength{\tabcolsep}{6pt} 
\renewcommand{\arraystretch}{1.2} 
\rowcolors{2}{gray!10}{white} 
\resizebox{\linewidth}{!}{%
\begin{tabular}{c l l c c c c c}
\toprule
\textbf{Training Size} & \textbf{Model} & \textbf{Method} & $\alpha$ & $\beta$ & $\gamma$ & \textbf{Learning Rate} & \textbf{Batch Size} \\ 
\midrule
5K  & Llama-3-Base & TPO/TPO-L        & 1    & 0.01      & \text{--}/0.5 & 5e-7      & 32  \\
5K  & Mistral-v0.3-Base & TPO/TPO-L      & 1/0.05    & 0.01/2      & \text{--}/1.6 & 5e-7      & 32  \\
10K & Llama-3-Base & TPO/TPO-L        & 1    & 0.01      & \text{--}/3   & 5e-7      & 32  \\
10K & Mistral-v0.3-Base & TPO/TPO-L     & 1/0.05    & 0.01/2      & \text{--}/1.6   & 5e-7      & 32  \\
20K & Llama-3-Base & TPO/TPO-L        & 1    & 0.01      & \text{--}/1.5 & 5e-7      & 128 \\
20K & Mistral-v0.3-Base & TPO/TPO-L        & 1    & 0.01/2      & \text{--}/1.6 & 5e-7      & 128 \\
40K & Llama-3-Base & TPO/TPO-L        & 1    & 0.01      & \text{--}/10  & 5e-7      & 64  \\
40K & Mistral-v0.3-Base & TPO/TPO-L        & 1/0.05    & 0.01/2      & \text{--}/1.6   & 5e-7      & 64  \\
60K & Llama-3-Instruct & TPO/TPO-L & 0.05 & 0.01/10   & \text{--}/3   & 1e-6      & 256 \\
60K & Mistral-v0.2-Instruct & TPO/TPO-L & 0.05 & 0.01/2.5   & \text{--}/0.3   & 1e-6      & 256 \\
\bottomrule
\end{tabular}
}
\end{table*}




%% file: tables/impact_ba_acc.tex
\begin{wraptable}{r}{.30\linewidth}
\small
\centering
\caption{Comparison of TPO on different hyperparameters on Arena-Hard.}
\label{tab:impact_ba_acc}
\resizebox{\linewidth}{!}{
\begin{tabular}{ccc}
\toprule

\multicolumn{0}{c}{\textbf{Method}} & \textbf{Scenario 1} & \textbf{Scenario 2} \\ 
\midrule
\textbf{$\text{TPO}_{y2}$} & 42.0 & 40.6 {\raisebox{-0.1em}{\scalebox{0.7}{\textcolor[HTML]{F62217}{\textbf{(-1.4)}}}}} \\
\textbf{$\text{TPO}_{y3}$} & 42.4 & 40.8 {\raisebox{-0.1em}{\scalebox{0.7}{\textcolor[HTML]{F62217}{\textbf{(-1.4)}}}}} \\
\textbf{$\text{TPO}_{y4}$} & 39.4 & 38.2 {\raisebox{-0.1em}{\scalebox{0.7}{\textcolor[HTML]{F62217}{\textbf{(-1.2)}}}}} \\

\bottomrule

\end{tabular}
}
\vspace{-2em}
\end{wraptable}

%% file: tables/mistral_instruction_benchmarks_instruct.tex
\begin{wraptable}{r}{.50\linewidth}
\vspace{-1.4em}
\small
\centering
\caption{Comparison of TPO and DPO on Instruction Benchmarks in the Instruct Setting for Mistral model.}
\label{tab:mistral_instruct_instruction_bench}
\resizebox{\linewidth}{!}{
\begin{tabular}{lcc}
\toprule

\multicolumn{1}{c}{\textbf{Method}} & \textbf{Arena-Hard} & \textbf{MixEval-Hard} \\

\midrule
\textbf{DPO} \footnotemark[1] & 19.9 & 36.5 \\
\textbf{$\text{TPO}_{y2}$} & 26.2 {\raisebox{-0.1em}{\scalebox{0.7}{\textcolor[HTML]{4CC417}{\textbf{(+6.3)}}}}} & 36.3 {\raisebox{-0.1em}{\scalebox{0.7}{\textcolor[HTML]{F62217}{\textbf{(-0.2)}}}}} \\
\textbf{$\text{TPO}_{y3}$} & 26.3 {\raisebox{-0.1em}{\scalebox{0.7}{\textcolor[HTML]{4CC417}{\textbf{(+6.4)}}}}} & 39.1 {\raisebox{-0.1em}{\scalebox{0.7}{\textcolor[HTML]{4CC417}{\textbf{(+2.6)}}}}}  \\
\textbf{$\text{TPO}_{y4}$} & 27.2 {\raisebox{-0.1em}{\scalebox{0.7}{\textcolor[HTML]{4CC417}{\textbf{(+7.3)}}}}} & 39.4 {\raisebox{-0.1em}{\scalebox{0.7}{\textcolor[HTML]{4CC417}{\textbf{(+2.9)}}}}}  \\


\bottomrule
\end{tabular}

}

\vspace{-0.5em}
\end{wraptable}

%% file: tables/mistral_instruction_benchmarks.tex
\setlength{\tabcolsep}{2pt}
\begin{table*}[!h]
\centering
\small 
\caption{Comparison of TPO and DPO on Instruction Benchmarks across the Base Settings for Mistral model.}

\resizebox{0.9\textwidth}{!}{
\begin{tabular}{l|cc|cc|cc|cc}
\toprule
\multirow{2}{*}{\textbf{Method}} & \multicolumn{2}{c}{\textbf{UltraFeedback (5k)}} & \multicolumn{2}{c}{\textbf{Ultrafeedback (10k)}} & \multicolumn{2}{c}{\textbf{UltraFeedback (20k)}} & \multicolumn{2}{c}{\textbf{UltraFeedback (40k)}}\\
& {\scriptsize \bf Arena-Hard} & {\scriptsize \bf MixEval-Hard} & {\scriptsize \bf Arena-Hard} & {\scriptsize \bf MixEval-Hard} & {\scriptsize \bf Arena-Hard} & {\scriptsize \bf MixEval-Hard} & {\scriptsize \bf Arena-Hard} & {\scriptsize \bf MixEval-Hard} \\
\midrule
\textbf{DPO} &  1.4 & 29.1 & 0.6 & 31.1 & 1.2 & 28.3 & 1.2 & 29.2 \\
\midrule
\textbf{TPO} & 4.3 {\raisebox{-0.1em}{\scalebox{0.7}{\textcolor[HTML]{4CC417}{\textbf{(+2.9)}}}}}  & 30.8 {\raisebox{-0.1em}{\scalebox{0.7}{\textcolor[HTML]{4CC417}{\textbf{(+1.7)}}}}}  & 6.2 {\raisebox{-0.1em}{\scalebox{0.7}{\textcolor[HTML]{4CC417}{\textbf{(+5.6)}}}}}  & 34.4 {\raisebox{-0.1em}{\scalebox{0.7}{\textcolor[HTML]{4CC417}{\textbf{(+3.3)}}}}}  & 4.5 {\raisebox{-0.1em}{\scalebox{0.7}{\textcolor[HTML]{4CC417}{\textbf{(+3.3)}}}}}  & 31.9 {\raisebox{-0.1em}{\scalebox{0.7}{\textcolor[HTML]{4CC417}{\textbf{(+3.6)}}}}}  & 7.4 {\raisebox{-0.1em}{\scalebox{0.7}{\textcolor[HTML]{4CC417}{\textbf{(+6.2)}}}}}  & 32.3 {\raisebox{-0.1em}{\scalebox{0.7}{\textcolor[HTML]{4CC417}{\textbf{(+3.1)}}}}}  \\
\bottomrule
\end{tabular}
}
\label{tab:mistral_instruct_res}
\end{table*}

\setlength{\tabcolsep}{6pt}

%% file: tables/mistral_different_chat_template.tex
\begin{table*}[!h]
\small
\centering
\caption{Comparison of TPO on different chat templates in Instruct setting for Mistral model.}
\label{tab:mistral_chat_template_analysis}
\resizebox{0.7\linewidth}{!}{%
\begin{tabular}{ccccc}
\toprule
\textbf{Method} &\textbf{Chat Template} & \textbf{GSM8K} & \textbf{Arena-Hard} & \textbf{MixEval-Hard} \\ 
\midrule
\textbf{$\text{TPO}_{y2}$} & Default & 40.6 & 26.2 & 36.3 \\
\textbf{$\text{TPO}_{y2}$} &  Pre-trained & 43.2 {\raisebox{-0.1em}{\scalebox{0.7}{\textcolor[HTML]{4CC417}{\textbf{(+2.6)}}}}} & 24.9 {\raisebox{-0.1em}{\scalebox{0.7}{\textcolor[HTML]{F62217}{\textbf{(-1.3)}}}}} & 38.9 {\raisebox{-0.1em}{\scalebox{0.7}{\textcolor[HTML]{4CC417}{\textbf{(+2.6)}}}}} \\

\bottomrule
\end{tabular}

}

\end{table*}

%% file: tables/mistral_downstream.tex
\begin{table}[h]
    \caption{Downstream task evaluation results of tasks on the huggingface open leaderboard and MMLU-Pro for mistral models. \label{tab:mistral_downstream_tpo}}
    \resizebox{\textwidth}{!}{\begin{tabular}{@{}lc|ccccccc@{}}
    \toprule
                   & \textbf{MMLU-Pro} & \textbf{MMLU (5)} & \textbf{ARC (25)} & \textbf{HellaSwag (10)} & \textbf{TruthfulQA (0)} & \textbf{Winograd (5)} & \textbf{GSM8K (5)} & \textbf{Average} \\ \midrule
                   \multicolumn{8}{c}{{\color[HTML]{222222} \textbf{Model: Mistral-7B - Data: UltraFeedback (5k)}}}                                                                                       \\ \midrule
                   \textbf{SFT}&21.12 & 51.03 & 48.54 & 79.36 &46.83&77.50&18.49 & 53.63\\
                   \textbf{DPO}&21.82 & 51.43 &49.31 &80.05&49.15&78.13&21.22 & 54.88\\
                   \textbf{CPO}&20.77 & 50.66 &48.80 &79.05&48.34&77.74&17.96 & 53.76\\
                   \textbf{IPO}&22.54 & 51.62 &50.42 &80.51&50.05&77.42&23.27 & 55.55\\
                   \textbf{ORPO}&20.97 & 50.32 &50.08 &79.93&46.27&77.82&18.87 & 53.88\\
                   \textbf{KTO}&21.68 & 51.42 &49.65 &80.08&49.18&77.82&20.92 & 54.85\\
                   \textbf{SimPO}&21.17 & 50.97 &48.89 &79.166&47.73&78.05&19.02 & 53.97 \\
                   \textbf{SLiC-HF}&20.58 & 50.64 &48.20 &79.04&48.31&77.90&18.80 & 53.81\\ \midrule
                   \textbf{TPO} & 32.86 &	62.29 & 60.15 & 83.26 & 53.36 & 78.77 & 42.00 & 63.30  \\ 
                   \textbf{TPO-L} & 34.52 & 61.72	& 62.88	& 84.46	& 51.30 & 79.16 & 42.15 & 63.61 \\  \midrule
                   \multicolumn{8}{c}{{\color[HTML]{222222} \textbf{Model: Mistral-7B - Data: UltraFeedback (10k)}}}                                                                            \\ \midrule
                   \textbf{SFT} & 20.50 & 49.33 & 49.31&78.14&47.57&75.61&16.90 & 52.81 \\
                   \textbf{DPO} & 21.69 & 49.83 &52.13&79.13&52.57&75.61&20.62 & 54.98 \\
                   \textbf{CPO} & 19.73 & 48.73 &48.63&77.68&50.08&76.08&17.66 & 53.14 \\
                   \textbf{IPO} & 22.00 & 49.80 &53.15&79.22&52.11&75.76&21.30 & 55.22\\
                   \textbf{ORPO} & 20.17 & 50.14 &50.42&79.04&47.56&75.61&17.81 & 53.43 \\
                   \textbf{KTO}  & 21.66  & 49.86 &51.96&79.20&52.68&75.45&21.07 & 55.04\\
                   \textbf{SimPO} & 20.65 & 49.60 &49.65&77.96&49.33&76.01&17.81 & 53.39 \\
                   \textbf{SLiC-HF} & 19.96 & 48.74&48.63&77.69&50.16&76.01&17.43 & 53.11 \\ \midrule
                   \textbf{TPO} & 32.43 &  61.87 & 61.17 &	83.30 &	53.64 &	79.08	& 41.17 & 63.37 \\ 
                   \textbf{TPO-L} & 35.49 & 61.57 & 65.78 & 85.18 & 61.32 & 79.71 & 42.46 & 66.00 \\  \midrule

                    \multicolumn{8}{c}{{\color[HTML]{222222} \textbf{Model: Mistral-7B - Data: UltraFeedback (20k)}}}                                                                            \\ \midrule
                   \textbf{SFT} & 24.8 & 55.31 &52.98&79.86 & 50.03&76.79&27.67 & 57.11 \\
                   \textbf{DPO} & 25.46 & 55.06 &55.97&80.36 &54.13&77.74&32.37 & 59.27 \\
                   \textbf{CPO} & 24.02 & 54.84 &53.24&79.64 &53.78&77.66&28.43 & 57.93 \\
                   \textbf{IPO} & 26.16 & 55.60 &56.99&80.04 &54.76&77.34&27.21 & 58.66 \\
                   \textbf{ORPO} & 25.19 & 55.39 &53.66&81.17 &51.06&77.66&29.34 & 58.05 \\
                   \textbf{KTO}   & 25.77 & 55.01 & 55.80 & 80.39 &53.99&77.66&31.61 & 59.08 \\
                   \textbf{SimPO}  & 25.48 & 55.37 &53.75&79.54 & 55.03&77.66&29.64 & 58.50\\
                   \textbf{SLiC-HF} & 24.11 & 54.72 &53.07&79.34&54.19&77.58&29.11 & 58.00 \\\midrule
                    \textbf{TPO} & 33.38 & 	62.17 &	61.34 &	82.92 &	55.64 &	79.24 &	44.20 & 64.25 \\ 
                   \textbf{TPO-L} &  36.17 &	62.24 &	65.10 &	84.83 &	61.29 &	80.03 &	46.55 & 66.67 \\  \midrule

                    \multicolumn{8}{c}{{\color[HTML]{222222} \textbf{Model: Mistral-7B - Data: UltraFeedback (40k)}}}                                                                            \\ \midrule
                   \textbf{SFT} & 23.90 &  54.62&50.93&78.72&47.73&77.74&28.80 & 56.42 \\
                   \textbf{DPO} & 24.58 & 53.85&51.96&79.06&52.78&77.42&30.62 & 57.62 \\
                   \textbf{CPO} & 23.08 & 54.25& 49.57& 78.48&51.54&77.34&27.89 & 56.51 \\
                   \textbf{IPO} & 25.44 & 53.63&54.18& 79.16&51.74& 77.34&28.20 & 57.38 \\
                   \textbf{ORPO}& 24.83 &53.97 & 52.21&80.75&50.39&77.26&11.67 & 54.38 \\
                   \textbf{KTO} & 24.83 & 53.80&52.30&79.23&51.64&77.58&30.85 & 57.57 \\
                   \textbf{SimPO} &24.68 & 54.26&52.38&78.60&53.05&77.82&31.15 & 57.88 \\
                   \textbf{SLiC-HF} & 23.08 & 54.32& 49.82&78.41&52.33&77.03&28.12 &56.67 \\ \midrule      
                   \textbf{TPO} &  32.03 &	61.32 &	60.58 &	82.40 &	53.41 &	78.61 &	39.20 & 62.58 \\ 
                   \textbf{TPO-L} & 36.17 & 60.18 & 65.10 & 84.83 & 61.29 & 80.03 & 32.68 & 65.80  \\           \midrule

                    \multicolumn{8}{c}{{\color[HTML]{222222} \textbf{Model: Mistral-7B-Instruct -  Data: Mistral-UltraFeedback-PairRM}}}                                                                                       \\ \midrule
                   \textbf{SFT} \footnotemark[1] & - & 60.4	& 63.57 & 	84.79 &	66.81 &	76.64 &	40.49 & 65.45 \\
                   \textbf{DPO} \footnotemark[1] & 33.02 &	60.53 &	65.36 &	85.86 &	66.71 &	76.8 &	40.33 & 65.93 \\
                   \textbf{CPO} \footnotemark[1] & 32.46 &	60.36 &	63.23 &	84.47 &	67.38 &	76.80 &	38.74 & 65.16 \\
                   \textbf{IPO} \footnotemark[1] & 32.87 &	60.20 &	63.31 &	84.88 &	67.36 &	75.85 &	39.42 & 65.17 \\
                   \textbf{ORPO} \footnotemark[1] & 32.41 &	60.43 &	61.43 &	84.32 &	66.33 &	76.80 &	36.85 & 64.36 \\
                    \textbf{KTO} \footnotemark[1] & 33.40 &	60.52 &	65.78 &	85.49 &	68.45 &	75.93 &	38.82 & 65.83  \\
                   \textbf{SimPO} \footnotemark[1] & 32.98 & 60.53 &	66.89 &	85.95 &	68.40 &	76.32 &	35.25 & 65.55 \\
                   \textbf{SLiC-HF} \footnotemark[1] & 32.75 &	60.59 &	59.90 &	84.05 &	65.30 &	76.32 &	39.65 & 64.30 \\ \midrule 
                   \textbf{$\text{TPO}_{y2}$} & 32.21 & 58.95 & 65.36 & 84.88 & 69.15 & 78.69 & 40.64 & 66.27 \\  
                   \textbf{$\text{TPO}_{y3}$} & 33.03 & 58.83 & 64.76 & 84.74 & 68.68 & 78.22 & 42.53 & 66.29  \\  
                   \textbf{$\text{TPO}_{y4}$} & 32.68 & 58.83 & 64.93 & 84.71 & 68.04 & 78.53 & 42.23 & 66.21 \\  
                   \textbf{$\text{TPO-L}_{y2}$} & 33.40 & 59.04 & 	65.10 &	84.95 &	67.95 &	78.30 &	41.85 & 66.19 \\

                   \bottomrule
                \end{tabular}}
\end{table}


%% file: tables/llama3_downstream.tex
\begin{table}[h]
    \caption{Downstream task evaluation results of tasks on the huggingface open leaderboard. \label{tab:downstream_tpo}}
    \resizebox{\textwidth}{!}{\begin{tabular}{@{}lccccccc@{}}
    \toprule
                   & \textbf{MMLU (5)} & \textbf{ARC (25)} & \textbf{HellaSwag (10)} & \textbf{TruthfulQA (0)} & \textbf{Winograd (5)} & \textbf{GSM8K (5)} & \textbf{Average} \\ \midrule
                   \multicolumn{8}{c}{{\color[HTML]{222222} \textbf{Model: Llama-3 Data: UltraFeedback (5k)}}}                                                                                       \\ \midrule
                   \textbf{SFT}   &    58.99        & 50.77            & 80.03                  & 47.12                  & 78.30                & 28.51             & 57.29           \\
                   \textbf{DPO} & 59.21 & 51.28 & 80.76 & 48.54 & 78.53 & 32.9 & 58.54 \\
                   \textbf{CPO}   & 58.97         & 50.77            & 80.28                  & 48.82                  & 78.06                & 31.92             & 58.14           \\
                   \textbf{IPO}   & 59.41            & 51.79            & 80.85                  & 48.99                  & 78.53                & 34.42             & 59.00           \\
                   \textbf{ORPO} & 59.27 & 52.22 & 80.65 & 47.09 & 77.43 & 33.97 & 58.44 \\
                   \textbf{KTO}   & 59.17            & 51.02            & 80.52                  & 48.13                  & 78.85                & 32.68             & 58.39          \\
                   \textbf{SimPO}  & 58.76           & 50.94            & 80.19                  & 47.87                  & 78.53                & 32.75             & 58.17           \\
                   \textbf{SLiC-HF} & 59.01           & 51.11            & 80.26                  & 48.83                  & 78.06                & 32.52            & 58.30           \\ \midrule
                   \textbf{TPO} & 65.34           & 59.04           & 81.90                  & 52.69                 & 77.11                & 51.86            & 64.66          \\ 
                   \textbf{TPO-L} & 65.33           & 59.04            & 82.51                 & 47.87                 & 78.06               & 52.08            & 64.15          \\  \midrule
                   \multicolumn{8}{c}{{\color[HTML]{222222} \textbf{Model: Llama-3 Data: UltraFeedback (10k)}}}                                                                            \\ \midrule
                   \textbf{SFT}   & 59.09           & 51.88            & 79.66                  & 42.80                  & 79.16                & 27.46             & 56.68           \\
                   \textbf{DPO} & 59.39 & 53.07 & 80.76 & 44.04 & 78.93 & 36.69 & 58.81 \\
               
                   \textbf{CPO}   & 59.32           & 51.71            & 79.46                  & 44.16                  & 79.79                & 32.83             & 57.88           \\
                   \textbf{IPO}   & 59.22            & 53.50            & 80.78                  & 43.81                  & 79.01                & 35.41             & 58.62           \\
                   \textbf{ORPO} & 59.69 & 52.99 & 81.03 & 42.82 & 78.53 & 36.16 & 58.54 \\
                   \textbf{KTO}   & 59.41            & 52.90            & 80.75                  & 43.92                  & 79.48                & 36.69             & 58.86           \\
                   \textbf{SimPO}  & 59.04            & 52.22            & 79.75                  & 42.99                  & 79.72                & 35.10             & 58.14           \\
                   \textbf{SLiC-HF} & 59.27            & 51.54            & 79.52                  & 44.11                  & 79.48                & 32.68            & 57.77    \\ \midrule
                   \textbf{TPO} & 65.35          & 59.39           & 81.96                 & 55.08                 & 77.51               & 52.16           & 65.24           \\ 
                   \textbf{TPO-L} & 65.43          & 61.01            & 83.07                  & 55.10                & 78.37                & 51.86           & 65.81          \\  \midrule

                    \multicolumn{8}{c}{{\color[HTML]{222222} \textbf{Model: Llama-3 Data: UltraFeedback (20k)}}}                                                                            \\ \midrule
                   \textbf{SFT}   & 62.51            &    52.82         & 80.46                  & 44.63                  & 77.11                & 20.32            &  56.31           \\
                   \textbf{DPO} &  62.85 & 55.03 & 82.19 & 47.48 & 77.19 & 47.99 & 62.12 \\
               
                   \textbf{CPO}   & 62.88            & 52.90            & 80.59                  & 49.08                  & 77.35                & 44.96             & 61.29           \\
                   \textbf{IPO}   & 62.79            & 57.42            & 82.03                  & 46.73                  & 77.19                & 45.49             & 61.94           \\
                   \textbf{ORPO} & 63.15 & 54.18 & 81.91 & 46.55 & 77.58 & 37.15 & 60.09 \\
                   \textbf{KTO}   & 62.74            & 55.12            & 82.23                  & 47.35                  & 77.19                & 48.67            & 62.22          \\
                   \textbf{SimPO}  & 62.98            & 54.35            & 80.81                  & 51.90                  & 78.45                & 45.03            & 62.25          \\
                   \textbf{SLiC-HF} & 62.90            &  53.33           & 80.62                  & 49.17                  & 77.66                & 44.12             &  61.30    \\\midrule
                    \textbf{TPO} & 65.30           & 59.81            & 81.75                  & 55.81                  & 77.19                & 52.99            & 65.48          \\ 
                   \textbf{TPO-L} & 65.27           & 60.15           & 82.75                & 54.86                  & 78.37                & 52.69          & 65.68           \\  \midrule

                    \multicolumn{8}{c}{{\color[HTML]{222222} \textbf{Model: Llama-3 Data: UltraFeedback (40k)}}}                                                                            \\ \midrule
                   \textbf{SFT}   & 62.19            & 53.92            & 80.14                  & 43.39                  & 77.98                & 39.20             & 59.47           \\
                   \textbf{DPO} &  61.84 & 56.74 & 80.73 & 48.59 & 77.51 & 45.11 & 61.75 \\
               
                   \textbf{CPO}   & 62.23            & 54.35            & 80.21                  & 46.31                  & 78.53                & 42.91             & 60.76           \\
                   \textbf{IPO}   & 62.09             & 58.11            & 81.54                  & 49.64                  & 77.74                &48.22             & 62.89           \\
                   \textbf{ORPO} & 62.36 & 56.31 & 82.25 & 45.62 & 78.45 & 40.94 & 60.99 \\
                   \textbf{KTO}   & 61.86            & 56.91            & 80.74                  & 48.55                  & 77.74                & 46.02            & 61.97           \\
                   \textbf{SimPO}  & 61.45            & 54.86            & 80.50                  & 47.73                  & 78.37                & 45.11             & 61.34           \\
                   \textbf{SLiC-HF} & 62.28            & 53.92            & 80.20                  & 46.38                  & 78.69                & 42.84             & 60.72    \\ \midrule      \textbf{TPO} & 64.85           & 58.96            & 81.53                  & 57.47                  & 77.66                & 51.18           & 65.28          \\ 
                   \textbf{TPO-L} & 65.15          & 64.93           & 79.40                & 63.43                  & 84.42                & 52.39            & 68.29         \\           \midrule

                    \multicolumn{8}{c}{{\color[HTML]{222222} \textbf{Model: Llama-3-Instruct Data: UltraFeedback-ArmoRM}}}                                                                                       \\ \midrule
                   \textbf{SFT} \footnotemark[1]   & 67.06            & 61.01            & 78.57                  & 51.66                  & 74.35                & 68.69             & 66.89           \\
                  
                   \textbf{DPO} \footnotemark[1]  & 66.88            & 63.99            & 80.78                  & 59.01                  & 74.66                & 49.81             & 65.86           \\
                   \textbf{CPO} \footnotemark[1] & 67.05 & 62.29 & 78.73 & 54.01 & 73.72 & 67.40 & 67.20 \\
                   \textbf{IPO}   & 66.52            & 61.95            & 77.90                  & 54.64                  & 73.09                & 58.23             & 65.39           \\
                   \textbf{ORPO} \footnotemark[1]  & 66.41            & 61.01            & 79.38                  & 54.37                  & 75.77                & 64.59             & 66.92           \\
                    \textbf{KTO} \footnotemark[1]  & 66.38            & 63.57            & 79.51                  & 58.15                  & 73.40                & 57.01             & 66.34           \\
                   \textbf{SimPO} \footnotemark[1] & 65.63            & 62.80            & 78.33                  & 60.70                  & 73.32                & 50.72             & 65.25           \\
                    \textbf{SLiC-HF} \footnotemark[1] & 66.41 & 61.26 & 78.80 & 53.23 & 76.16 & 66.57 & 67.07 \\ \midrule 
                    
                   \textbf{$\text{TPO}_{y2}$} & 65.89 & 66.38                  & 79.38                 & 59.53 & 74.59                & 77.18 & 70.49          \\ 
                   \textbf{$\text{TPO}_{y3}$} & 65.54 & 65.61           &     79.22            &   58.75                & 75.00               & 77.71           & 70.30         \\    
                   \textbf{$\text{TPO}_{y4}$} & 65.69 & 64.33           & 79.10                & 55.17                 & 75.30                & 78.09            & 69.61         \\    
                   \textbf{$\text{TPO-L}_{y2}$} & 65.68 & 66.04           & 79.37                & 58.66                  & 75.30                & 77.26            & 70.39         \\

                   \bottomrule
                \end{tabular}}
\end{table}
